\documentclass[runningheads]{llncs}
\usepackage[T1]{fontenc}
\usepackage{graphicx}
\usepackage{subcaption}
\usepackage{multicol}
\usepackage{dcolumn}

\usepackage{rotating}

\newcolumntype{.}{D{.}{.}{-1}}

\usepackage{url}

\usepackage{amsmath}
\usepackage{authblk}

\begin{document}
\title{SnapE -- Training Snapshot Ensembles\\of Link Prediction Models}
%
%
\author{Ali Shaban\inst{1,2}\thanks{This work is not related to the position at Amazon.} \and Heiko Paulheim\inst{1}\orcidID{0000-0003-4386-8195}}
\institute{University of Mannheim, Data and Web Science Group, Germany\\
\email{heiko.paulheim@uni-mannheim.de} \and
Amazon Web Services, USA\\\email{shaban-a@live.com}}

%
%
\authorrunning{Ali Shaban and Heiko Paulheim}

\maketitle              
\begin{abstract}
Snapshot ensembles have been widely used in various fields of prediction. They allow for training an ensemble of prediction models at the cost of training a single one. They are known to yield more robust predictions by creating a set of diverse base models.
In this paper, we introduce an approach to transfer the idea of snapshot ensembles to link prediction models in knowledge graphs. Moreover, since link prediction in knowledge graphs is a setup without explicit negative examples, we propose a novel training loop that iteratively creates negative examples using previous snapshot models. An evaluation with four base models across four datasets shows that this approach constantly outperforms the single model approach, while keeping the training time constant.

\end{abstract}

\section{Introduction}
Knowledge Graphs (KGs) are a crucial instrument for structuring complex, interrelated data in a way that is interpretable both by humans and machines~\cite{heist2020knowledge,hogan2021knowledge}. They are used in many applications ranging from search engines to recommendation systems and from natural language processing to computer vision.

KGs often exhibit sparseness and incompleteness \cite{Dong14knowledgevault,Paulheim17kgrefinment}. This link sparseness can significantly hinder the KGs' potential use in applications like semantic search. As a result, filling the missing links has gained traction in the AI community. This task is called \emph{Link Prediction}, and it is one of the primary applications of machine learning and embedding methods on Knowledge Graphs \cite{bordes13transe}. Various research efforts have validated the efficacy of link prediction in addressing the sparsity issue in KGs and reinforcing their potential \cite{nickel15relational}.
 
Link prediction with knowledge graph embeddings (KGEs) faces many challenges that have been the subject of research~\cite{biswas2023knowledge}, including the sparsity of graphs~\cite{Mahmudur18classimbalance}, noisy or erroneous data~\cite{Zhang16noisykg}, and scalability~\cite{Duan16scale}, among others. In machine learning, many of those challenges are addressed by ensemble methods. Ensembles can improve generalization by reducing the overfitting often seen in individual predictors \cite{zhou2012ensemble}, and better handle incomplete or noisy data by leveraging the robustness resulting from the combination of multiple models that make errors on different instances \cite{Polikar06ensembleconsensus}. Finally, ensemble methods can be used to address the scalability challenges of link predictions by training different models on different sub-graphs and combining them \cite{Duan16scale} or by training smaller models on the whole graphs and combining them into an ensemble~\cite{xu21lowdim}. The approach followed in this paper falls into the latter category.

Snapshot Ensembles are an effective way of training ensembles without increasing the training time or computation cost~\cite{huang17snapshot}. By storing snapshots of a single model at distinct points in its training process and combining them into an ensemble, they require the same training time as a single model~\cite{huang17snapshot}. These snapshots, stored at the end of each cycle during training with cyclic learning rate schedules, have diverse model behaviors, ensuring less correlation among base models and avoiding the model being stuck in local minima. In many fields, snapshot ensembles have been shown to outperform single base models.

In this paper, we show how to apply the idea of snapshot ensembles to link prediction in KGs. We train an ensemble of base models in the same time as we would classically train a single model, and show that the ensemble outperforms the single base model in many cases. Moreover, we propose a new technique of negative sampling, where the previous snapshots are used to generate adversarial samples for the next training cycle.
The contributions of this paper are as follows:
\begin{itemize}
    \item We propose SnapE, a new \emph{training method} which can be applied to existing link prediction models.\footnote{Note: we do \emph{not} propose a new link prediction model, but a new training method for existing models.}
    \item We show that using the same resources for training (measured by training time or memory), we can improve the performance of existing link prediction models.
    \item We propose a new negative sampling method which can be used with SnapE.
\end{itemize}

The rest of this paper is structured as follows. We provide a brief overview of ensemble methods for link prediction in knowledge graphs in section~\ref{sec:related}. We sketch our approach in section~\ref{sec:approach}, followed by an evaluation and an ablation study in section~\ref{sec:evaluation}. We conclude with a summary and an outlook on future work.

\section{Related Work}
\label{sec:related}
Link prediction in knowledge graphs is an extensively researched area. Approaches can be roughly categorized as geometric, matrix factorization, and neural network based approaches~\cite{rossi2021knowledge,wang2021survey}.

One of the earliest approaches to use ensembles for link prediction has been proposed by Krompass et al. \cite{Krompass15kge}. They train three different heterogeneous models (i.e., a TransE, a RESCAL, and an mwNN model), and make a prediction by averaging the probabilities for each predicted triple. Similar approaches are pursued by Rivas-Barragan et al. \cite{Barragan22ekgem} and Gregucci et al. \cite{Gregucci23attention}. They can improve the prediction quality, but also signficantly increase the training efforts.

Prabhakara et al. \cite{Prabhakar2023scale} divide a knowledge graph into separate subgraphs, and train an embedding model on each subgraph. The predictions of each single embedding model are aggregated using the min, max, or mean of the individual scores. The results are mixed, and no evidence of an actual performance improvement is given, i.e., there is no evidence that training $m$ models on a smaller subgraph is computationally cheaper than training one model on the full graph. Wan et al. \cite{Wan2020} also consider the training on subgraphs, but instead of combining predictions, they combine the embedding vectors of entities and relations. While they can show that the resulting model is more robust to noise in the knowledge graph, their paper also lacks a discussion of training cost.

The approach closest to ours is Xu et al. \cite{xu21lowdim}, which proposes an approach of training different models of the same base class at a lower dimensionality independently, initialized with different random seeds, and combining their scores. They show that training the ensemble of different models with the same memory footprint exposes a better prediction quality than a single large model. Unlike our approach, they train the set of different models in in multiple parallel runs, rather than a single training run (i.e., Xu et al. train $m$ models in $N$ epochs \emph{each}, which requires a total of $N \times m$ epochs, whereas we train $m$ models in $N$ epochs \emph{in total}). Since this approach is the closest to SnapE, we compare our work to theirs, also empirically.

While most of the approaches above can show an improvement in the link prediction quality, they usually require a larger training effort, e.g., by training more than one full model. In contrast, SnapE, as proposed in this paper, is the \textbf{first approach to train an ensemble of link prediction models at the same training cost of one single full model}.

\section{Approach}
\label{sec:approach}
\begin{figure}[t]
  \centering
  \begin{subfigure}[t]{0.49\columnwidth}
    \centering
    \includegraphics[width=\textwidth]{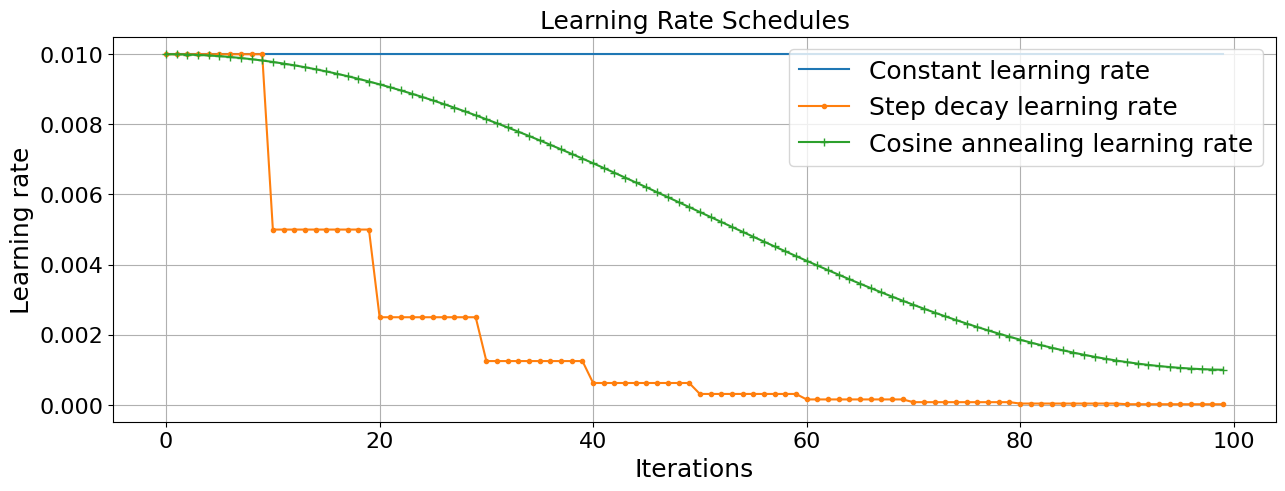}
    \caption{Non-cyclic}
    \label{fig:learning-rate-non-cyclic}
  \end{subfigure}
  \begin{subfigure}[t]{0.49\columnwidth}
    \centering
    \includegraphics[width=\textwidth]{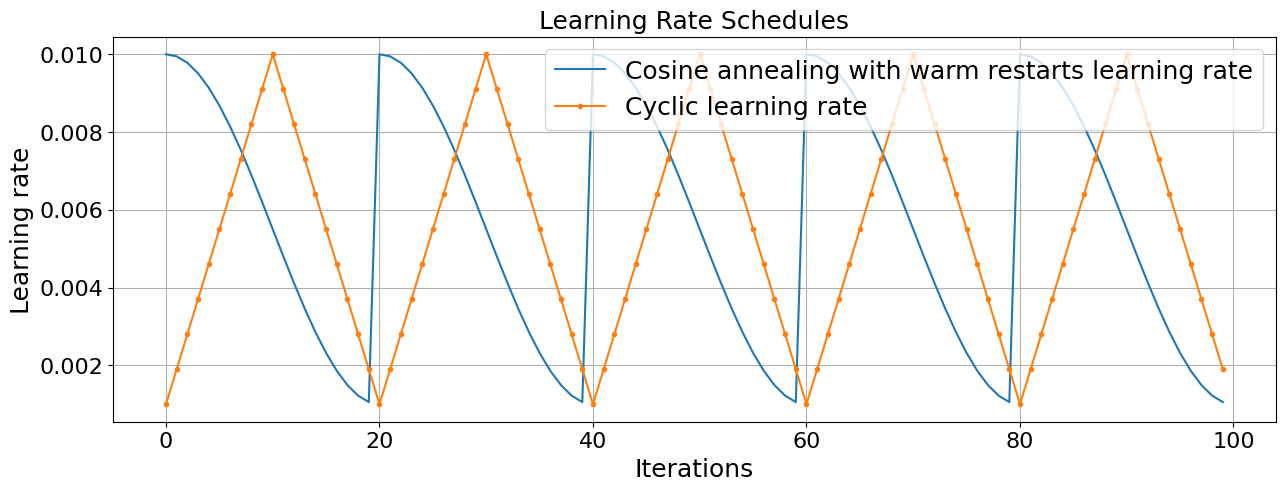}
    \caption{Cyclic}
    \label{fig:learning-rate-cyclic}
  \end{subfigure}
  \caption{Non-cyclic vs. cyclic learning rate schedules}
  \label{fig:lr_schedules}
\end{figure}

KGE models are trained by iteratively adapting their weights in order to maximize a target function. In each iteration, the model's weights are adapted so that the error function is lowered. The amount by which each weight can be changed in each iteration is called the \emph{learning rate}. A high learning rate leads to a fast convergence of the model, 
but may not yield the best possible solution. A low learning rate, on the other hand, requires more time for convergence, but may yield better results. 
Therefore, many models are trained with a decaying learning rate, as shown in Fig.~\ref{fig:learning-rate-non-cyclic}. There are different decay functions for learning rates, such as step decay~\cite{Ge19stepdecay} or cosine annealing~\cite{loshchilov17sgdr}.

The idea of Snapshot Ensembles, introduced by Huang et al. in 2017 \cite{huang17snapshot}, uses a cyclic learning rate, as shown in Fig.~\ref{fig:learning-rate-cyclic}. Each time the learning rate schedule reaches a local minimum, a snapshot of the model is stored. At prediction time, each stored model is used to predict, and the predictions are combined. The original snapshot ensemble paper uses cyclic cosine annealing, as shown in Fig.~\ref{fig:learning-rate-cyclic}, and averages the predictions. In this paper, we use the same mechanism for training an ensemble of link prediction models. We experiment with different learning rate schedulers, score aggregation functions, and negative samplers.

\subsection{Learning Rate Schedulers}
As discussed above, the original snapshot ensemble uses the cosine cyclic annealing schema (CCA), defined as~\cite{loshchilov17sgdr}:
\begin{equation}
\alpha(t) = \frac{\alpha_0}{2} \left(\cos\left((\frac{\pi mod(t-1,\lceil T/M \rceil)}{\lceil T/M \rceil}\right) + 1\right)
\label{eqn:cosine_annealing}
\end{equation}
where $\alpha_0$ is the initial maximum learning rate, \textit{t} is the iteration number, \textit{T} is the total number of training iterations, and \textit{M} is the number of cycles. A later paper used a variation called max-min cyclic cosine learning rate (MMCCLR), defined as~\cite{wen19normlr}:
\begin{equation}
\eta_t = \eta_{\text{min}} + \frac{1}{2} \left( \eta_{\text{max}} - \eta_{\text{min}} \right) \left( 1 + \cos\left(\frac{mod(\lceil t/b\rceil,\lceil T/Mb \rceil)}{\lceil T/Mb \rceil} \pi \right) \right)
\label{eqn:MMCCLR}
\end{equation}
where $\eta_t$ is the learning rate at step t, $\eta_{min}$ and $\eta_{max}$ are the minimum and maximum learning rates, respectively, and $b$ is the batch size.
The difference to cyclic cosine annealing is that the learning rate does not get as close to $0$, which may lead to local minima being exploited a little better in the case of CCA.

In addition to those two schedulers, we explore two further variants, called \emph{deferred} cyclic annealing. Assuming that a link prediction model requires some initial training in order to learn the principal structure of a knowledge graph, we defer the cyclic annealing by using a constant learning rate for the first $k$ epochs before starting the cyclic annealing. The first model is then stored when the learning rate hits the first minimum. The result is a smaller number of models in a fixed set of epochs, but, on the other hand, those models may be a better fit due to the longer warmup time of the snapshot model.

\begin{figure}[t!]
  \centering
  \begin{subfigure}[t]{0.49\columnwidth}
    \centering
    \includegraphics[width=\textwidth]{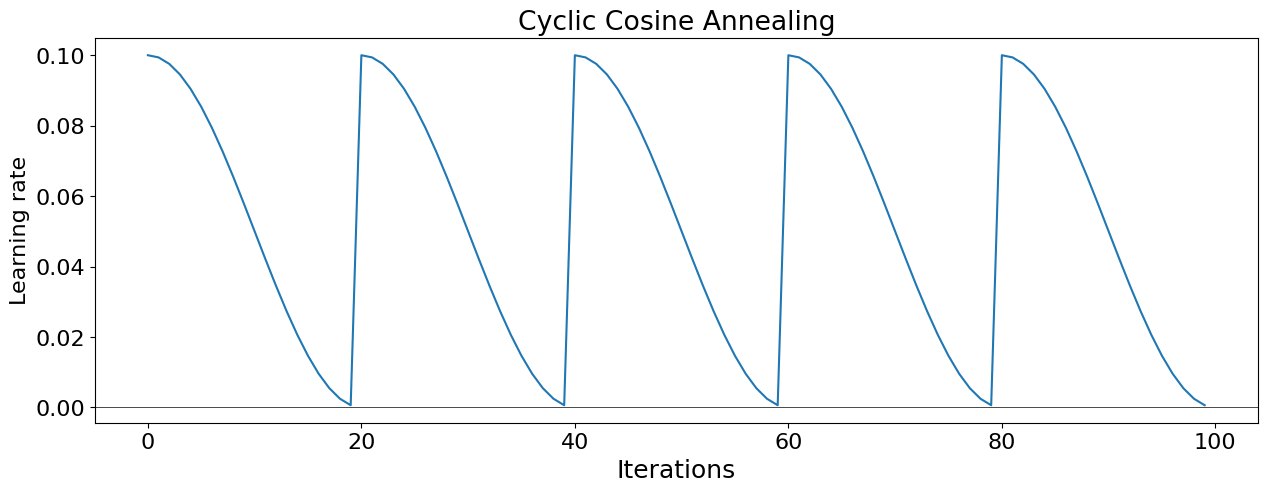}
    \label{fig:cyclic-cosine-annealing}
  \end{subfigure}
  \begin{subfigure}[t]{0.49\columnwidth}
    \centering
    \includegraphics[width=\textwidth]{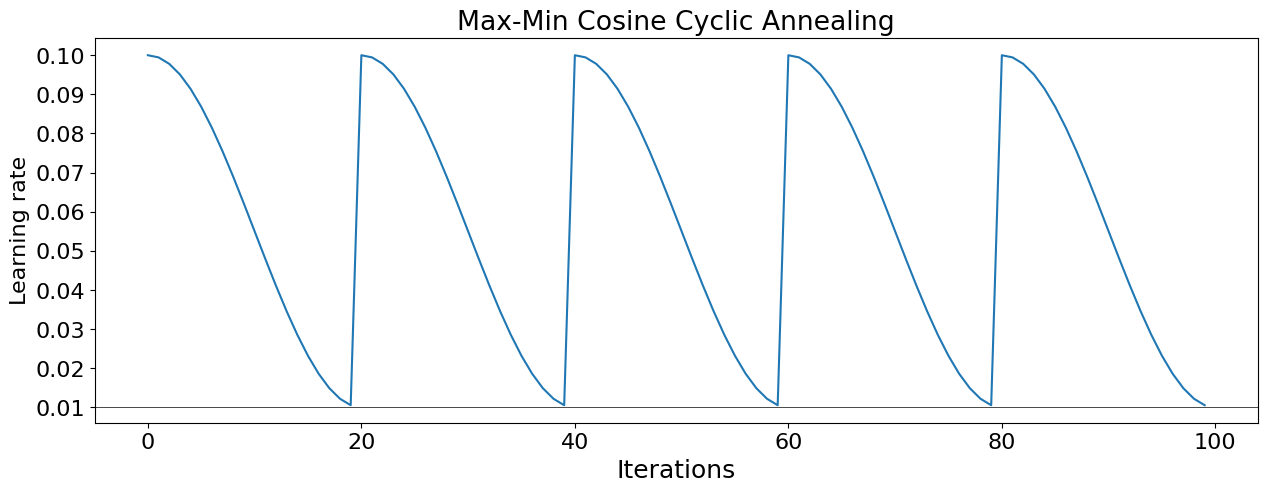}
    \label{fig:max-min-cosine-annealing}
  \end{subfigure}
    \begin{subfigure}[t]{0.49\columnwidth}
    \centering
    \includegraphics[width=\textwidth]{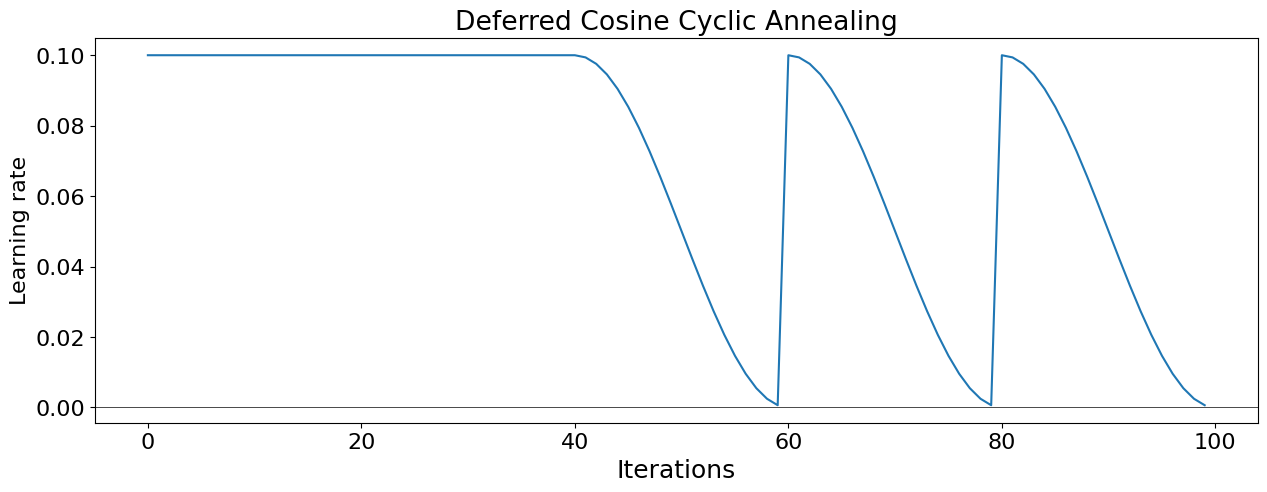}
    \label{fig:deferred-cyclic-cosine-annealing}
  \end{subfigure}
  \begin{subfigure}[t]{0.49\columnwidth}
    \centering
    \includegraphics[width=\textwidth]{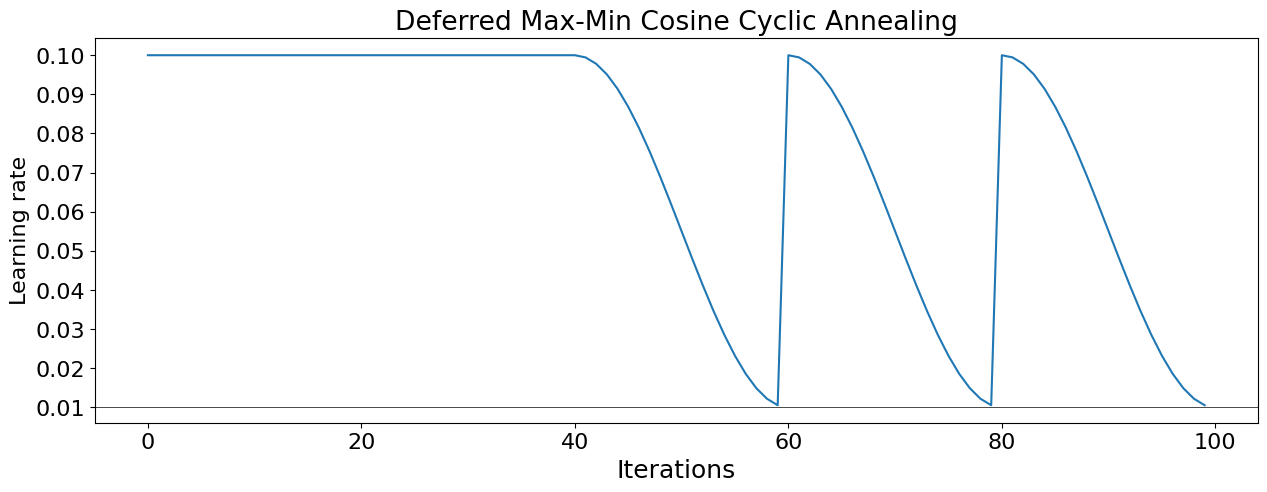}
    \label{fig:deferred-max-min-cosine-annealing}
  \end{subfigure}
  \caption{Learning rate schedules used in this paper. Snapshots are always stored at the minima of the learning rate schedules.}
  \label{fig:lr_schedules_paper}
\end{figure}

Fig.~\ref{fig:lr_schedules_paper} illustrates the learning rate schedulers used in this paper.

\subsection{Combining Predictions}
In order to come up with a final prediction from the individual base models, their scores need to be aggregated. We follow different strategies:
\begin{description}
\item[Simple average:]{For each prediction made with the base models, we normalize the scores of each model with a min-max scaler (so that each model's scores fall into a $[0,1]$ range), and average those normalized scores.}
\item[Weighted average:]{Instead of using a simple average, we weigh each model by (a) a weight derived from the model's last training loss\footnote{$max + min - loss$, where $max$ and $min$ are the maximum and minimum loss of the models in the ensemble}, or (b) with descending weights from the last stored base model, assigning the highest weight to the last model, assuming that this is the best fit.}
\item[Borda rank aggregation:]{The rankings are unified using the Borda rank algorithm~\cite{dwork01rank} (without considering the scores).}
\end{description}

\subsection{Negative Samplers}
Unlike many other machine learning problems, link prediction in knowledge graphs only comes with positive examples. Therefore, link prediction requires the creation of negative examples, which is usually done via randomly corrupting positive examples, i.e., for a given triple $<s,p,o>$, either $s$ or $o$ is replaced with a random other entity to form a corrupted triple $<s',p,o>$ or $<s,p,o'>$, respectively.

In addition to this standard sampling, we propose an extended negative sampler, which uses the last snapshot to create negative examples for the next cycle. The idea is similar to boosting~\cite{FREUND97boosting}, where an ensemble of models is trained in a sequence, with each model focusing on the mistakes made by the previous ones.

After storing a snapshot model, the model is used to predict o' or s' for a triple $<s,p,o>$ in the training set. The highest scoring negative $<s,p,o'>$ or $<s',p,o>$ is then added as a negative example. With that approach, we guide the optimization to learn more diverse models, where subsequent models correct mistakes made by models stored in previous iterations. 

\begin{figure}[t]
  \centering
  \begin{subfigure}[t]{0.49\columnwidth}
    \centering
    \includegraphics[width=\textwidth]{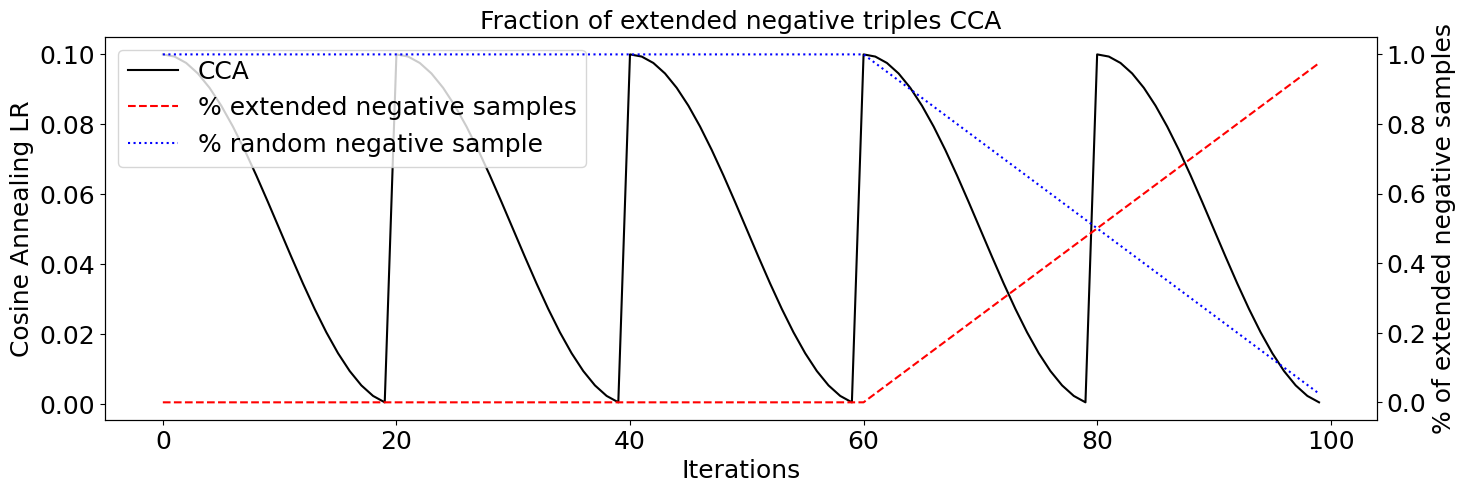}
    \label{fig:ext_neg_cca}
  \end{subfigure}
  \begin{subfigure}[t]{0.49\columnwidth}
    \centering
    \includegraphics[width=\textwidth]{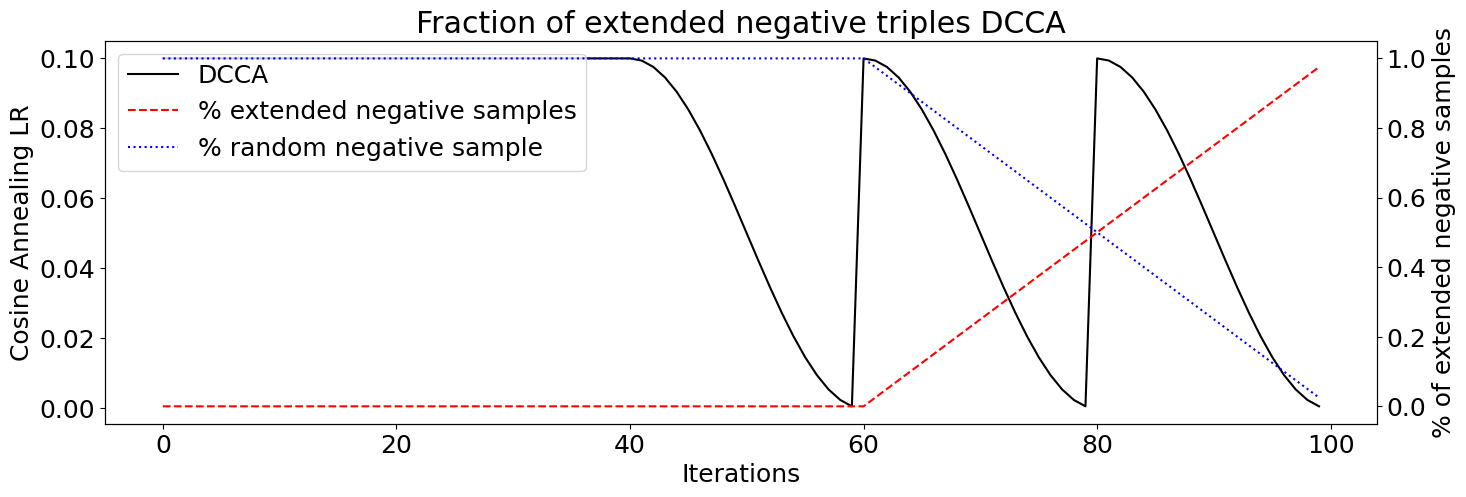}
    \label{fig:ext_neg_dcca}
  \end{subfigure}
  \caption{Usage of the extended negative sampler in the cosine annealing (left) and deferred cosine annealing (right) setting.}
  \label{fig:ext_neg}
\end{figure}

Like for the deferred cyclic annealing variants, we want to make sure that the models first learn the basic information in the graph before focusing on the edge cases. Therefore, we use random negative samples in the first iterations and successively use the extended sampling mechanism in the later stages. Fig.~\ref{fig:ext_neg} illustrates this idea.

\begin{figure*}[t]
    \centering
    \includegraphics[width=\textwidth]{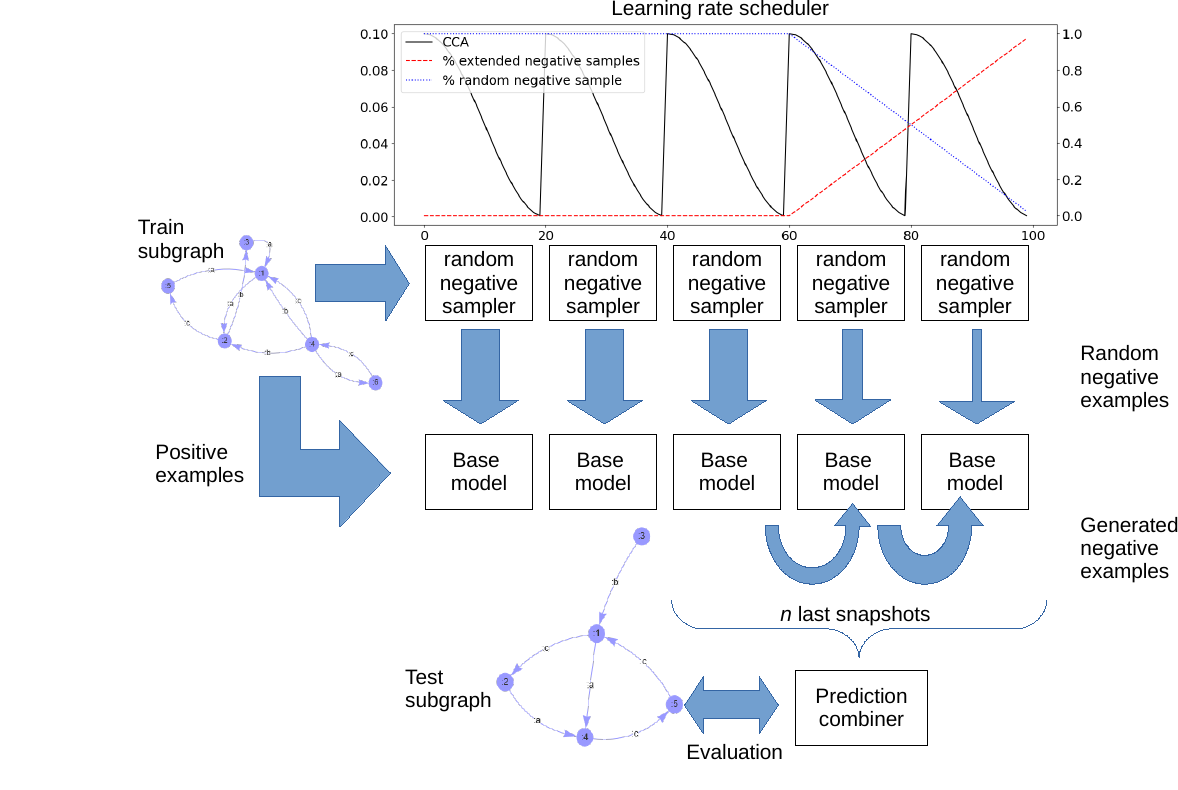}
    \caption{Illustration of the overall approach}
    \label{fig:overall_approach}
\end{figure*}

Fig.~\ref{fig:overall_approach} illustrates the big picture. Following a learning rate schedule, the approach trains a base model and stores several snapshots. The last $n$ snapshots are used for forming the final ensemble and computing a combined prediction. It is possible to use all snapshots, but since the later ones are expected to have a higher quality, we foresee the possibility to also only combine the latest ones. The negative samples in the first rounds are generated randomly; in later iterations, they can be successively replaced by the extended sampler, which uses the (wrong) predictions of the previous snapshots as non-trivial negative examples.

\section{Evaluation}
\label{sec:evaluation}
We run experiments with our approach and four different base models and compare them to the results achieved with the respective baseline models. In our evaluation, we consider two different setups:
\begin{enumerate}
    \item \textbf{Equal training time budget.} We train a model with $d$ dimensions and $m$ snapshots, and compare it to a single base model with $d$ dimensions.
    \item \textbf{Equal memory budget.} We train a model with $\frac{d}{m}$ dimensions and $m$ snapshots, and compare it to a single base model with $d$ dimensions.
\end{enumerate}
The code for reproducing the results is available online.
\footnote{\url{https://github.com/Alishaba/pykeen-snapshot-ensembles}}

We compare our approach to the one proposed by Xu et al.~\cite{xu21lowdim}, who also train an ensemble of lower dimensional models, coined as M\emph{base}, i.e., MTransE, MDistMult, MComplEx, and MRotatE. In order to ensure comparability to our approach, we use the same number of epochs for those models, train a total of 10 models (which is the maximum number of models for SnapE in our experiments,see section~\ref{subsec:evaluation_parameters}), and stick to the parameters in the original paper otherwise (i.e., a learning rate of 0.0003, Adam optimizer, and random negative sampling).\footnote{Note that the setup is pretty similar to SnapE. The only differences are that each model is initialized randomly, that the sampling is always random, i.e., never uses a previous model to generate negative samples, and that the combination uses uniform weights. Therefore, and due to the implementation of Xu et al.~\cite{xu21lowdim} not being available, we use the SnapE implementation to emulate MTransE etc. This also allows for a fair comparison of training and prediction times.} All experiments were run on a server with 24 NVIDIA A100 GPUs with 32 GB of RAM each.


\subsection{Datasets}
We evaluate our approach on four different datasets, i.e., DBpedia50~\cite{Shi17dbpedia}, FB15k-237~\cite{Toutanova15fb15k}, WN18RR~\cite{Dettmers17wn18rr}, and AristoV4~\cite{chen2021relation}. The characteristics of the four datasets are depicted in table~\ref{tab:datasets}.\footnote{We use the training, test, and validation split as provided in PyKEEN~\cite{ali21pykeen}.}

\begin{table}[t]
    \caption{Datasets used for the Evaluation}
    \label{tab:datasets}
    \scriptsize
    \centering
    \begin{tabular}{l|r|r|r}
         & Entities & Relations & Triples \\
         \hline
         DBpedia50 & 24,624 & 351 & 34,421 \\
         FB15k237 & 14,505 & 237 & 310,079 \\
         WN18RR & 40,559 & 11 & 92,584 \\
         AristoV4 & 42,016 & 1,593 & 279,425 \\
    \end{tabular}
\end{table}
\begin{sidewaystable}
    \caption{Results of SnapE compared to baselines and to the M\emph{base} ensemble approach. Ensemble approaches outperforming their representative baselines are marked in bold. The best overall results are underlined. The last column shows the average change in HITS@10 over the respective base model across all four datasets. Moreover, we show the average change of each metric for each dataset across all models.}
    \label{tab:main_results}
    \centering
    \scriptsize
\begin{tabular}{l|r|r|r||r|r|r||r|r|r||r|r|r||r}
                  & \multicolumn{3}{c||}{DBpedia50}                                          & \multicolumn{3}{c||}{FB15k237}                                           & \multicolumn{3}{c||}{WN18RR}                                             & \multicolumn{3}{c||}{AristoV4}                                           & \multicolumn{1}{c}{avg.}  \\
Model             & H@1                    & H@10                   & MRR                    & H@1                    & H@10                   & MRR                    & H@1                    & H@10                   & MRR                    & H@1                    & H@10                   & MRR                    & $\Delta$HITS@10           \\ 
\hline
\multicolumn{14}{c}{Baselines}                                                                                                                                                                                                                                                                                                                            \\ 
\hline
TransE$_d=64$     & 0.068                  & 0.276                  & 0.150                  & 0.116                  & 0.418                  & 0.216                  & 0.005                  & 0.319                  & 0.123                  & 0.067                  & 0.221                  & 0.120                  & –                         \\
TransE$_d=128$    & 0.056                  & 0.245                  & 0.125                  & 0.097                  & 0.427                  & 0.206                  & 0.002                  & 0.301                  & 0.112                  & 0.037                  & 0.224                  & 0.102                  & –                         \\
DistMult$_d=64$   & 0.262                  & 0.378                  & 0.309                  & 0.090                  & 0.318                  & 0.167                  & 0.292                  & 0.309                  & 0.299                  & 0.047                  & 0.151                  & 0.084                  & –                         \\
DistMult$_d=128$  & 0.299                  & 0.418                  & 0.345                  & 0.105                  & 0.343                  & 0.185                  & 0.293                  & 0.328                  & 0.308                  & 0.060                  & 0.170                  & 0.098                  & –                         \\
ComplEx$_d=64$    & 0.191                  & 0.319                  & 0.239                  & 0.091                  & 0.416                  & 0.197                  & 0.271                  & 0.381                  & 0.313                  & 0.038                  & 0.130                  & 0.067                  & –                         \\
ComplEx$_d=128$   & 0.002                  & 0.008                  & 0.005                  & 0.094                  & 0.402                  & 0.195                  & 0.282                  & 0.360                  & 0.311                  & 0.064                  & 0.176                  & 0.105                  & –                         \\
RotatE$_d=64$     & 0.223                  & 0.269                  & 0.242                  & 0.167                  & 0.455                  & 0.265                  & 0.315                  & 0.348                  & 0.328                  & 0.089                  & 0.220                  & 0.135                  & –                         \\
RotatE$_d=128$    & 0.227                  & 0.288                  & 0.250                  & 0.178                  & 0.483                  & 0.281                  & 0.328                  & 0.379                  & 0.346                  & 0.038                  & 0.130                  & 0.069                  & –                         \\ 
\hline
\hline
\multicolumn{14}{c}{SnapE – Same Memory Budget}                                                                                                                                                                                                                                                                                                           \\ 
\hline
TransE$_d=64$     & 0.062                  & 0.250                  & 0.131                  & 0.108                  & 0.395                  & 0.202                  & 0.003                  & 0.289                  & 0.106                  & 0.055                  & 0.184                  & 0.100                  & -10.23\%                  \\
TransE$_d=128$    & \textbf{0.058}         & 0.243                  & 0.124                  & \textbf{0.128}         & \textbf{0.445}         & \textbf{0.231}         & \textbf{0.004}         & 0.289                  & 0.110                  & \textbf{0.067}         & \textbf{0.226}         & \textbf{0.120}         & +0.05\%                   \\
DistMult$_d=64$   & 0.256                  & 0.351                  & 0.292                  & \textbf{0.097}         & 0.299                  & 0.165                  & 0.276                  & 0.305                  & 0.288                  & 0.045                  & 0.145                  & 0.080                  & -4.52\%                   \\
DistMult$_d=128$  & 0.263                  & 0.394                  & 0.313                  & \textbf{0.138}         & \textbf{0.355}         & \textbf{0.211}         & 0.291                  & 0.319                  & 0.303                  & \textbf{0.060}         & \textbf{0.175}         & \textbf{0.101}         & -0.49\%                   \\
ComplEx$_d=64$    & \textbf{0.266}         & \textbf{0.349}         & \textbf{0.296}         & \textbf{0.150}         & \textbf{0.479}         & \textbf{0.258}         & \textbf{0.321}         & \textbf{0.391}         & \textbf{0.347}         & \textbf{0.056}         & \textbf{0.174}         & \textbf{0.097}         & +15.29\%                  \\
ComplEx$_d=128$   & \textbf{0.006}         & \textbf{0.029}         & \textbf{0.014}         & \textbf{0.171}         & \textbf{0.512}         & \textbf{0.283}         & \textbf{0.318}         & \textbf{0.389}         & \textbf{0.344}         & 0.060                  & 0.172                  & 0.098                  & +75.44\%                  \\
RotatE$_d=64$     & 0.208                  & 0.261                  & 0.227                  & 0.148                  & 0.431                  & 0.243                  & 0.310                  & 0.337                  & 0.319                  & 0.064                  & 0.190                  & 0.107                  & -6.25\%                   \\
RotatE$_d=128$    & \textbf{0.227}         & \textbf{0.289}         & 0.250                  & 0.174                  & 0.470                  & 0.274                  & 0.318                  & 0.357                  & 0.332                  & \textbf{0.086}         & \textbf{0.204}         & \textbf{0.127}         & +12.21\%                  \\ 
\hline
avg. $\Delta$     & +28.02\%               & +31.57\%               & +22.25\%               & +24.63\%               & +3.85\%                & +10.53\%               & +8.93\%                & -1.94\%                & -0.78\%                & +24.64\%               & +7.27\%                & +12.57\%               &                           \\ 
\hline
\hline
\multicolumn{14}{c}{SnapE – Same Training Time Budget}                                                                                                                                                                                                                                                                                                    \\ 
\hline
TransE$_d=64$     & \textbf{0.141}         & \textbf{0.348}         & \textbf{0.223}         & \textbf{0.135}         & \textbf{0.427}         & \textbf{0.231}         & 0.005                  & 0.313                  & 0.122                  & \textbf{0.076}         & \textbf{0.244}         & \textbf{0.133}         & +9.26\%                   \\
TransE$_d=128$    & \textbf{0.086}         & \textbf{0.284}         & \textbf{0.161}         & \textbf{0.140}         & \textbf{0.457}         & \textbf{0.246}         & \textbf{0.004}         & 0.299                  & 0.111                  & \textbf{0.082}         & \textbf{\underline{0.258}} & \textbf{0.142}         & +9.38\%                   \\
DistMult$_d=64$   & \textbf{0.288}         & \textbf{0.395}         & \textbf{0.327}         & \textbf{0.141}         & \textbf{0.354}         & \textbf{0.214}         & \textbf{0.303}         & \textbf{0.338}         & \textbf{0.317}         & \textbf{0.059}         & \textbf{0.179}         & \textbf{0.102}         & +10.94\%                  \\
DistMult$_d=128$  & \textbf{\underline{0.326}} & \textbf{0.433}         & \textbf{\underline{0.367}} & \textbf{0.146}         & \textbf{0.366}         & \textbf{0.221}         & \textbf{0.299}         & \textbf{0.346}         & \textbf{0.316}         & \textbf{0.064}         & 0.170                  & \textbf{0.102}         & +3.86\%                   \\
ComplEx$_d=64$    & \textbf{0.250}         & \textbf{0.334}         & \textbf{0.280}         & \textbf{0.221}         & \textbf{0.543}         & \textbf{0.329}         & \textbf{0.323}         & \textbf{\underline{0.403}} & \textbf{0.352}         & \textbf{0.095}         & \textbf{0.232}         & \textbf{0.143}         & +29.90\%                  \\
ComplEx$_d=128$   & \textbf{0.030}         & \textbf{0.068}         & \textbf{0.044}         & \textbf{0.198}         & \textbf{0.518}         & \textbf{0.305}         & \textbf{0.312}         & \textbf{0.367}         & \textbf{0.332}         & \textbf{0.072}         & \textbf{0.181}         & \textbf{0.110}         & +194.82\%                 \\
RotatE$_d=64$     & \textbf{0.256}         & \textbf{0.321}         & \textbf{0.280}         & \textbf{0.167}         & 0.453                  & 0.262                  & \textbf{0.323}         & \textbf{0.364}         & \textbf{0.339}         & \textbf{0.084}         & \textbf{0.224}         & 0.132                  & +6.31\%                   \\
RotatE$_d=128$    & \textbf{0.264}         & \textbf{0.336}         & \textbf{0.290}         & \textbf{0.180}         & 0.481                  & \textbf{0.282}         & \textbf{0.335}         & \textbf{0.388}         & \textbf{0.353}         & \textbf{\underline{0.100}} & \textbf{0.246}         & \textbf{\underline{0.149}} & +26.94\%                  \\ 
\hline
avg. $\Delta$     & +207.64\%              & +104.56\%              & +115.13\%              & +51.50\%               & +10.72\%               & +24.46\%               & +14.64\%               & +3.36\%                & +3.97\%                & +61.12\%               & +27.07\%               & +38.49\%               &                           \\ 
\hline
\hline
\multicolumn{14}{c}{M\textit{base}}                                                                                                                                                                                                                                                                                                                       \\ 
\hline
MTransE$_d=64$    & 0.016                  & \textbf{\underline{0.487}} & \textbf{0.214}         & \textbf{0.184}         & \textbf{0.558}         & \textbf{0.314}         & 0.000                  & \textbf{0.368}         & \textbf{0.154}         & \textbf{0.078}         & \textbf{0.340}         & \textbf{0.169}         & +44.74\%                  \\
MTransE$_d=128$   & 0.013                  & \textbf{0.479}         & \textbf{0.210}         & \textbf{0.113}         & \textbf{0.582}         & \textbf{0.283}         & \textbf{0.003}         & \textbf{0.349}         & \textbf{0.164}         & 0.014                  & \textbf{\underline{0.360}} & \textbf{0.142}         & +52.13\%                  \\
MDistMult$_d=64$  & \textbf{0.301}         & \textbf{0.381}         & \textbf{0.332}         & \textbf{0.095}         & 0.259                  & 0.149                  & 0.028                  & 0.028                  & 0.028                  & 0.041                  & 0.116                  & 0.068                  & -33.04\%                  \\
MDistMult$_d=128$ & \textbf{\underline{0.326}} & 0.326                  & 0.293                  & 0.090                  & 0.304                  & 0.165                  & 0.241                  & 0.279                  & 0.265                  & \textbf{0.072}         & \textbf{0.224}         & \textbf{0.124}         & -4.09\%                   \\
MComplEx$_d=64$   & 0.005                  & 0.028                  & 0.013                  & \textbf{0.243}         & \textbf{0.592}         & \textbf{0.361}         & 0.009                  & 0.035                  & 0.018                  & \textbf{0.051}         & \textbf{0.153}         & \textbf{0.088}         & -30.55\%                  \\
MComplEx$_d=128$  & 0.000                  & 0.001                  & 0.001                  & \textbf{0.197}         & \textbf{0.522}         & \textbf{0.310}         & 0.266                  & 0.268                  & 0.267                  & 0.043                  & 0.115                  & 0.043                  & -29.53\%                  \\
MRotatE$_d=64$    & 0.195                  & 0.255                  & 0.219                  & \textbf{0.234}         & \textbf{0.594}         & \textbf{0.357}         & \textbf{0.325}         & \textbf{0.384}         & \textbf{0.344}         & \textbf{\underline{0.133}} & \textbf{0.311}         & \textbf{\underline{0.193}} & +19.27\%                  \\
MRotatE$_d=128$   & \textbf{0.228}         & \textbf{0.344}         & \textbf{0.268}         & \underline{\textbf{0.236}} & \underline{\textbf{0.615}} & \underline{\textbf{0.366}} & \underline{\textbf{0.437}} & 0.354                  & \underline{\textbf{0.382}} & \textbf{0.122}         & \textbf{0.277}         & \textbf{0.174}         & +38.28\%                  \\ 
\hline
avg. $\Delta$     & -42.35\%               & -1.72\%                & -9.23\%                & +51.96\%               & +21.24\%               & +33.54\%               & -28.01\%               & -23.35\%               & -15.76\%               & +29.14\%               & +32.48\%               & +32.00\%               &

\end{tabular}
\end{sidewaystable}

\subsection{Base Models}
As base models, we chose TransE~\cite{bordes13transe}, DistMult~\cite{yang2015distmult}, ComplEx~\cite{trouillon16complex}, and RotatE~\cite{sun19rotate}. For each base model, we train two baseline models, one with 64 and one with 128 dimensions, and a batch size of 128 each. We determine the optimal number of epochs by applying early stopping, using $relative\_delta = 0.000001$ and $patience = 2$, and $Hits@10$ as a target metric. Furthermore, we pick the best learning rate $lr \in \{10, 1, 0.1, 0.01\}$ and the best optimizer out of stochastic gradient descent (SGD), Adam, and AdaGrad by applying grid search. The number of epochs that worked best for each base model is then also used for the snapshot ensemble (while the learning rate is controlled by the respective decay scheme).\footnote{For all other parameters, we use the default values from PyKEEN, including negative sampling rate (1 negative sample per positive example) and loss function (MarginRankingLoss for TransE, DistMult, and Rotate, SoftplusLoss for ComplEx).}
The parameters used for each model are listed in the appendix.

Note that the goal is not to achieve the best possible baseline results~\cite{ruffinelli2019you}. We are more interested in analyzing the performance gains that can be obtained by using snapshot ensembles, compared to the standard training loop.

\subsection{Parameters}
\label{subsec:evaluation_parameters}
For the parameters of the snapshot ensemble, we use the number of episodes that worked best on for the baseline model on the respective dataset (see table~\ref{tab:base_parameters} in the appendix). Moreover, we use scores weighted by last training loss as an aggregation method, min-max-scaling to normalize the scores before, and deferred CCA as a scheduler. We pick the no. of cycles $C$ out of $\{5,10\}$ and the number of models $M$ to be used out of $\left[2,C\right]$ ((cf. Fig.~\ref{fig:lr_schedules_paper})\footnote{Note that when $M=C$, then there is actually no warmup phase, i.e., the schedule is equal to a non-deferred one.}, as well as the best performing optimizer. The parameters used for each model are listed in the appendix.

\subsection{Results}
Table~\ref{tab:main_results} shows a comparison of SnapE against the baselines. We show the results both for the setup with the same memory budget as well as for the setup with the same time budget.

We can observe that training with SnapE outperforms the standard training mechanism in about a third of the cases when considering the same memory budget case, and in almost all cases when considering the same training time budget case (in that case at the price of a larger overall model). Moreover, in all cases, the best overall result is achieved by an ensemble model.

It can also be observed that not all models benefit equally from the SnapE training. The gains for RotatE are moderate, while the other models, especially DistMult and ComplEx, benefit much more strongly. In particular, ComplEx (and, to a lesser extent, DistMult) strongly benefit on FB15k237 and WN18RR, whereas the improvements for TransE and RotatE are not that strong on those two datasets. As stated in the original RotatE paper, composition patterns are very important in those two datasets, and ComplEx and DistMult are not capable of learning those patterns. So one possible explanation is that the ensembling compensates for the inability of those models to learn composition patterns. 

Comparing to the M\emph{base} models proposed by \cite{xu21lowdim}, we can observe that while they sometimes achieve the best overall performance, their average gains are less (and often even negative) than those of SnapE. Hence, we conclude that while SnapE is occasionally outperformed by M\emph{base}, it is more stable overall. Given the observations about the runtime below, we observe that SnapE provides a better trade-off of computational efforts and prediction quality than M\emph{base}.

\subsection{Runtime Behavior}
\begin{table*}[t!]
    \caption{Training and prediction time (s) of the baselines, the SnapE-trained approaches, and the M\emph{base} ensembles}
    \label{tab:runtime}
    \scriptsize
    \centering
    \begin{tabular}{l|r|r||r|r||r|r||r|r}
         & \multicolumn{2}{c||}{DBpedia50} & \multicolumn{2}{c||}{FB15k237} & \multicolumn{2}{c||}{WN18RR} & \multicolumn{2}{c}{AristoV4} \\
         Model & train & predict & train & predict & train & predict & train & predict\\
         \hline
         \multicolumn{9}{c}{Baselines}\\
         \hline
         TransE$_{d=64}$ & 58.427 & 0.190 & 358.516 & 1.466 & 303.580 & 0.656 & 645.440 & 1.994\\
         TransE$_{d=128}$ & 48.723 & 0.316 & 1,687.463 & 2.134 & 128.592 & 0.517 & 599.860 & 2.831\\
         DistMult$_{d=64}$ & 84.571 & 0.340 & 196.292 & 1.490 & 49.885 & 0.411 & 186.774 & 2.417\\
         DistMult$_{d=128}$ & 22.309 & 0.234 & 185.398 & 1.626 & 77.129 & 0.517 & 1,690.012 & 2.768\\
         ComplEx$_{d=64}$ & 256.092 & 0.333 & 2,643.319 & 1.706 & 370.614 & 0.401 & 1,463.573 & 1.956\\
         ComplEx$_{d=128}$ & 29.171 & 0.372 & 1,319.236 & 1.805 & 344.264 & 0.566 & 11,334.286 & 5.792\\
         RotatE$_{d=64}$ & 62.807 & 0.241 & 1,130.800 & 1.620 & 1,109.552 & 0.532 & 1,962.201 & 2.764\\
         RotatE$_{d=128}$ & 92.876 & 0.393 & 1,687.463 & 2.135 & 169.264 & 0.743 & 1,677.724 & 4.825\\
        \hline
        \hline
         \multicolumn{9}{c}{SnapE -- Same Memory Budget}\\
         \hline
        TransE$_{d=64}$	&	54.006	&	0.203	&	296.375	&	1.314	&	275.402	&	0.649	&	578.863	&	1.878\\
        TransE$_{d=128}$	&	38.658	&	0.236	&	443.577	&	1.503	&	103.982	&	0.755	&	619.845	&	2.258\\
        DistMult$_{d=64}$	&	55.419	&	0.203	&	126.954	&	1.638	&	31.698	&	0.648	&	182.741	&	1.781\\
        DistMult$_{d=128}$	&	13.182	&	0.239	&	258.389	&	1.599	&	49.283	&	0.783	&	198.814	&	2.474\\
        ComplEx$_{d=64}$	&	180.451	&	0.185	&	2,470.871	&	1.252	&	322.798	&	0.544	&	1,050.948	&	1.569\\
        ComplEx$_{d=128}$	&	23.597	&	0.282	&	836.381	&	1.518	&	300.829	&	0.738	&	2,243.378	&	2.129\\
        RotatE$_{d=64}$	&	56.665	&	0.219	&	1,119.917	&	1.452	&	99.450	&	0.784	&	1,336.407	&	2.317\\
        RotatE$_{d=128}$	&	66.058	&	0.362	&	1,594.601	&	1.689	&	128.422	&	1.143	&	1,482.312	&	3.306\\
        \hline
        avg. $\Delta$ & -23.86\%	&	-17.77\%	&	-17.07\%	&	-13.18\%	&	-30.21\%	&	+40.17\%	&	-31.16\%	&	-24.21\%\\
        \hline
         \multicolumn{9}{c}{SnapE -- Same Training Time Budget}\\
        \hline
        \hline
        TransE$_{d=64}$	&	91.298	&	0.637	&	397.608	&	2.374	&	291.219	&	0.782	&	930.551	&	5.007\\
        TransE$_{d=128}$	&	68.966	&	0.871	&	653.377	&	3.213	&	118.345	&	1.106	&	1,202.238	&	10.263\\
        DistMult$_{d=64}$	&	14.589	&	0.293	&	153.939	&	3.491	&	57.423	&	2.426	&	256.140	&	7.571\\
        DistMult$_{d=128}$	&	18.579	&	0.672	&	294.357	&	4.472	&	60.385	&	1.506	&	1,784.194	&	3.241\\
        ComplEx$_{d=64}$	&	200.428	&	0.269	&	842.800	&	2.881	&	329.754	&	0.734	&	1,652.348	&	6.809\\
        ComplEx$_{d=128}$	&	34.277	&	1.069	&	925.995	&	3.910	&	306.544	&	1.094	&	2,410.949	&	3.123\\
        RotatE$_{d=64}$	&	80.620	&	1.034	&	149.655	&	4.392	&	223.709	&	4.574	&	1,733.061	&	5.713\\
        RotatE$_{d=128}$	&	120.118	&	1.825	&	1,752.696	&	2.535	&	613.024	&	7.561	&	1,772.830	&	5.028\\
        \hline
        avg. $\Delta$ & +6.47\%	&	+180.78\%	&	-24.25\%	&	+99.65\%	&	+17.71\%	&	+333.58\%	&	+14.43\%	&	+119.61\%\\
        \hline
         \multicolumn{9}{c}{M\emph{base}}\\
        \hline
        \hline
MTransE$_d=64$    & 547.341   & 1.070               & 2,856.178  & 3.301             & 2,773.676 & 3.598            & 5,496.94   & 8.27             \\
MTransE$_d=128$   & 424.415   & 1.615               & 4,343.727  & 5.108             & 1,061.422 & 4.234            & 5,078.27   & 12.73            \\
MDistMult$_d=64$  & 547.341   & 1.070               & 903.491    & 3.290             & 195.154   & 2.558            & 836.39     & 7.31             \\
MDistMult$_d=128$ & 167.593   & 0.981               & 965.005    & 5.410             & 516.893   & 4.229            & 15,772.17  & 13.29            \\
MComplEx$_d=64$   & 1,925.209 & 1.217               & 23,684.887 & 3.681             & 962.539   & 2.319            & 12,888.71  & 7.55             \\
MComplEx$_d=128$  & 147.732   & 1.058               & 11,655.125 & 5.773             & 2,895.896 & 5.013            & 10,317.45  & 12.48            \\
MRotatE$_d=64$    & 589.339   & 1.674               & 11,047.856 & 5.361             & 1,029.588 & 4.473            & 18,374.42  & 13.92            \\
MRotatE$_d=128$   & 713.364   & 2.362               & 16,925.643 & 9.354             & 1,287.857 & 7.531            & 6,434.62   & 22.24            \\ 
\hline
avg. $\Delta$     & +671.36\% & +369.21\%           & +624.30\%  & +190.34\%         & +494.37\% & +665.77\%        & +571.36\%  & +301.59\%       

    \end{tabular}
\end{table*}
\begin{sidewaystable}
    \caption{Analysis of the variance of models in the SnapE ensembles. The table contrasts the HITS@10 scores achieved by SnapE and by the best single model in the ensemble (the better of the two marked in bold). The column N-\# shows which is the best performing single model (0 means the last, 1 means the second to last, etc.). Moreover, we depict the average pairwise correlation of the models in the ensemble. For M\emph{base}, we do not report the number of the best performing model, since all models are independent.}
    \label{tab:variance_analysis}
    \center 
    \scriptsize
\begin{tabular}{l|r|r|r|r||r|r|r|r||r|r|r|r||r|r|r|r}
                 & \multicolumn{4}{c||}{DBpedia50}                           & \multicolumn{4}{c||}{FB15k237}                             & \multicolumn{4}{c||}{WN18RR}                               & \multicolumn{4}{c}{AristoV4}                                \\
                 & Ensemble       & \multicolumn{2}{r|}{Best Single} & avg.  & Ensemble       & \multicolumn{2}{r|}{Best Single} & avg.   & Ensemble       & \multicolumn{2}{r|}{Best Single} & avg.   & Ensemble       & \multicolumn{2}{r|}{Best Single} & avg.    \\
Model            & H@10           & H@10           & N-\#            & corr. & H@10           & H@10  & N-\#                     & corr.  & H@10           & H@10           & N-\#            & corr.  & H@10           & H@10  & N-\#                     & corr.   \\ 
\hline
\multicolumn{17}{c}{SnapE – Same Memory Budget}                                                                                                                                                                                                                      \\ 
\hline
TransE$_d=64$    & 0.250          & \textbf{0.252} & 0               & 0.324 & \textbf{0.395} & 0.390 & 1                        & 0.484  & \textbf{0.289} & 0.289          & 1               & -0.235 & \textbf{0.184} & 0.179 & 1                        & 0.628   \\
TransE$_d=128$   & \textbf{0.243} & 0.243          & 1               & 0.632 & \textbf{0.445} & 0.440 & 5                        & 0.408  & 0.289          & \textbf{0.290} & 1               & -0.211 & \textbf{0.226} & 0.217 & 1                        & 0.512   \\
DistMult$_d=64$  & \textbf{0.351} & 0.336          & 1               & 0.962 & \textbf{0.299} & 0.119 & 4                        & 0.222  & 0.305          & \textbf{0.306} & 1               & 0.996  & \textbf{0.145} & 0.066 & 3                        & 0.274   \\
DistMult$_d=128$ & \textbf{0.394} & 0.389          & 1               & 0.652 & \textbf{0.355} & 0.260 & 7                        & 0.105  & 0.319          & \textbf{0.322} & 1               & 0.991  & \textbf{0.175} & 0.112 & 3                        & 0.178   \\
ComplEx$_d=64$   & \textbf{0.349} & 0.337          & 1               & 0.005 & \textbf{0.479} & 0.462 & 2                        & -0.007 & \textbf{0.391} & 0.386          & 1               & -0.308 & \textbf{0.174} & 0.137 & 1                        & 0.349   \\
ComplEx$_d=128$  & \textbf{0.029} & 0.005          & 0               & 0.393 & \textbf{0.512} & 0.419 & 2                        & 0.439  & \textbf{0.389} & 0.384          & 1               & -0.224 & \textbf{0.172} & 0.162 & 1                        & -0.079  \\
RotatE$_d=64$    & \textbf{0.261} & 0.226          & 0               & 0.133 & \textbf{0.431} & 0.428 & 1                        & 0.189  & 0.337          & \textbf{0.337} & 1               & 0.228  & \textbf{0.190} & 0.180 & 1                        & -0.245  \\
RotatE$_d=128$   & \textbf{0.289} & 0.259          & 1               & 0.037 & \textbf{0.470} & 0.467 & 6                        & 0.158  & 0.357          & 0.357          & 1               & 0.079  & \textbf{0.204} & 0.203 & 1                        & -0.204  \\ 
\hline
\hline
\multicolumn{17}{c}{SnapE – Same Training Time Budget}                  \\ 
\hline
TransE$_d=64$    & \textbf{0.348} & 0.345          & 1               & 0.416 & \textbf{0.427} & 0.381 & 5                        & 0.719  & 0.313          & \textbf{0.313} & 1               & -0.110 & \textbf{0.244} & 0.206 & 7                        & 0.620   \\
TransE$_d=128$   & \textbf{0.284} & 0.231          & 0               & 0.837 & \textbf{0.457} & 0.401 & 5                        & 0.837  & \textbf{0.299} & 0.299          & 1               & 0.085  & \textbf{0.258} & 0.175 & 3                        & 0.504   \\
DistMult$_d=64$  & \textbf{0.395} & 0.365          & 1               & 0.489 & \textbf{0.354} & 0.288 & 8                        & 0.153  & \textbf{0.338} & 0.281          & 7               & 0.219  & \textbf{0.179} & 0.116 & 9                        & 0.323   \\
DistMult$_d=128$ & \textbf{0.433} & 0.401          & 4               & 0.375 & \textbf{0.366} & 0.338 & 8                        & 0.101  & \textbf{0.346} & 0.333          & 2               & 0.967  & \textbf{0.170} & 0.167 & 1                        & 0.743   \\
ComplEx$_d=64$   & \textbf{0.334} & 0.291          & 2               & 0.397 & \textbf{0.543} & 0.383 & 7                        & 0.638  & \textbf{0.403} & 0.398          & 1               & -0.236 & \textbf{0.232} & 0.088 & 1                        & 0.325   \\
ComplEx$_d=128$  & \textbf{0.068} & 0.008          & 0               & 0.264 & \textbf{0.518} & 0.400 & 6                        & 0.654  & \textbf{0.367} & 0.358          & 1               & 0.033  & \textbf{0.181} & 0.176 & 1                        & -0.100  \\
RotatE$_d=64$    & \textbf{0.321} & 0.263          & 1               & 0.085 & \textbf{0.453} & 0.344 & 8                        & 0.102  & \textbf{0.364} & 0.277          & 0               & 0.006  & \textbf{0.224} & 0.205 & 3                        & 0.000   \\
RotatE$_d=128$   & \textbf{0.336} & 0.264          & 2               & 0.089 & \textbf{0.481} & 0.476 & 1                        & 0.224  & \textbf{0.388} & 0.314          & 9               & 0.003  & \textbf{0.246} & 0.245 & 1                        & -0.260 \\
\hline
\hline
\multicolumn{17}{c}{M\emph{base}}                  \\ 
\hline
MTransE$_d=64$    & \textbf{0.487} & 0.129          &       --        & 0.571                     & \textbf{0.558}                 & 0.393 &       --                 & 0.871                     & \textbf{0.368} & 0.274          &       --        & 0.716                     & \textbf{0.340} & 0.195 &       --                 & 0.880                       \\
MTransE$_d=128$   & \textbf{0.479} & 0.140          &       --        & 0.494                     & \textbf{0.582}                 & 0.422 &       --                 & 0.840                     & \textbf{0.349} & 0.184          &       --        & 0.686                     & \textbf{0.360} & 0.211 &       --                 & 0.716                       \\
MDistMult$_d=64$  & \textbf{0.381} & 0.306          &       --        & 0.936                     & \textbf{0.259}                 & 0.212 &       --                 & 0.954                     & \textbf{0.028} & 0.021          &       --        & 0.979                     & \textbf{0.116} & 0.075 &       --                 & 0.969                       \\
MDistMult$_d=128$ & \textbf{0.326} & 0.258          &       --        & 0.981                     & \textbf{0.304}                 & 0.292 &       --                 & 0.931                     & \textbf{0.279} & 0.183          &       --        & 0.962                     & \textbf{0.224} & 0.204 &       --                 & 0.879                       \\
MComplEx$_d=64$   & \textbf{0.028} & 0.002          &       --        & 0.551                     & \textbf{0.592}                 & 0.404 &       --                 & 0.658                     & \textbf{0.035} & 0.001          &       --        & 0.474                     & \textbf{0.153} & 0.070 &       --                 & 0.827                       \\
MComplEx$_d=128$  & 0.001          & 0.001          &       --        & 0.007                     & \textbf{0.522}                 & 0.260 &       --                 & 0.586                     & \textbf{0.268} & 0.003          &       --        & 0.561                     & \textbf{0.115} & 0.052 &       --                 & 0.705                       \\
MRotatE$_d=64$    & \textbf{0.255} & 0.228          &       --        & 0.540                     & \textbf{0.594}                 & 0.518 &       --                 & 0.755                     & \textbf{0.384} & 0.349          &       --        & 0.417                     & \textbf{0.311} & 0.240 &       --                 & 0.797                       \\
MRotatE$_d=128$   & \textbf{0.344} & 0.286          &       --        & 0.286                     & \textbf{0.615}                 & 0.529 &       --                 & 0.738                     & \textbf{0.437} & 0.373          &       --        & 0.303                     & \textbf{0.277} & 0.205 &       --                 & 0.827                      

\end{tabular}
\end{sidewaystable}

Table~\ref{tab:runtime} shows the runtime behavior of the baselines as well as the SnapE configurations for the same memory and the same training time budget.\footnote{While it might seem surprising that the larger baseline models with d=128 often train faster than the smaller ones with d=64, despite having a larger number of parameters, this is due to the early stopping criterion, which allowed the larger models to be trained in a smaller number of episodes (see appendix), leading to overall lower training times.}

In both cases, we can observe that the training time is similar to or even smaller than the one for the baseline method.

On the other hand, the prediction time is significantly larger for the same training time budget setup. The reason is that instead of making one model predict, \emph{each} of the snapshot models has to make a prediction, which leads to the higher prediction time. The combination of the individual predictions also contributes to the prediction time, but to a lesser extent. For the same memory setup, this effect is not observed, since a prediction with a number of smaller models is not more time-consuming than a prediction with one large model.

For the M\emph{base} approach, we observe significantly larger training times\footnote{However, the training in M\emph{base} can be parallelized, which is not possible for SnapE.}, as already discussed in section~\ref{sec:related}, since SnapE trains all its models in the same number of epochs that M\emph{base} uses per model. The fact that the prediction time of M\emph{base} is also higher is due to the fact that M\emph{base} always uses all base models trained, while in SnapE, this is a hyperparameter, and often a smaller number of models is used (see appendix). 

\subsection{Model Variety}
One essential prerequisite for using ensemble methods successfully is that the base models expose some variety. Generally, the more diverse the single models in an ensemble are, the better the ensemble.

To measure the model variety, we look at the triple scores produced by the individual models, and compute their correlation for each pair of models. Table~\ref{tab:variance_analysis} shows the average of those correlations for all models in the ensemble, together with the HITS@10 of the best single base model and the ensemble.

We can see that the approach is indeed effective, since in most cases, the ensembles do not only outperform the baseline, but also the single best model.\footnote{Otherwise, one could have assumed that SnapE performs better simply due to a smaller model size and hence, e.g., a lesser tendency to overfit.} Exceptions are only observed for the same memory setup for WN18RR, where in half of the cases, the best individual model is slightly better than the ensemble. Moreover, the gains over the best single model are often more considerable in cases of less correlated models, i.e., when SnapE learns more diverse base models.

What is also important is that the best performing model is only rarely the last one -- otherwise, the overhead of storing all intermediate models would be even more questionable. Moreover, also in the cases where the best single model is found early on in the training process (indicated by a high value in the N-\# column in table~\ref{tab:variance_analysis}), there are often considerable performance gains compared to the best single model, meaning that the subsequent models are not better individually, but are able to capture information and cases not solved correctly by the best single model.
Interestingly, negative correlations are also observed (i.e., SnapE learns \emph{contradicting} models). In many of those cases, we can see that the ensemble still performs well.

In comparison to M\emph{base}, we observe that SnapE most often produces less correlated (i.e., more diverse) models, and, at the same, also better single models, which are then often learned late in the training process.

\subsection{Ablation Study}
We analyze the impact of (1) different optimizers, (2) different learning rate schedulers (see Fig.~\ref{fig:lr_schedules_paper}), (3) different negative samplers, (4) different strategies for combining the predictions of the snapshots.
In all the ablation studies, we use the best performing parameter sets, and vary only the respective parameter of study. For the sake of space, we show only the impact on Hits@10, and we only consider the setting with the same training time budget.

\begin{figure*}[t]
    \centering
    \begin{subfigure}[t]{0.49\textwidth}
        \centering
        \includegraphics[width=\textwidth]{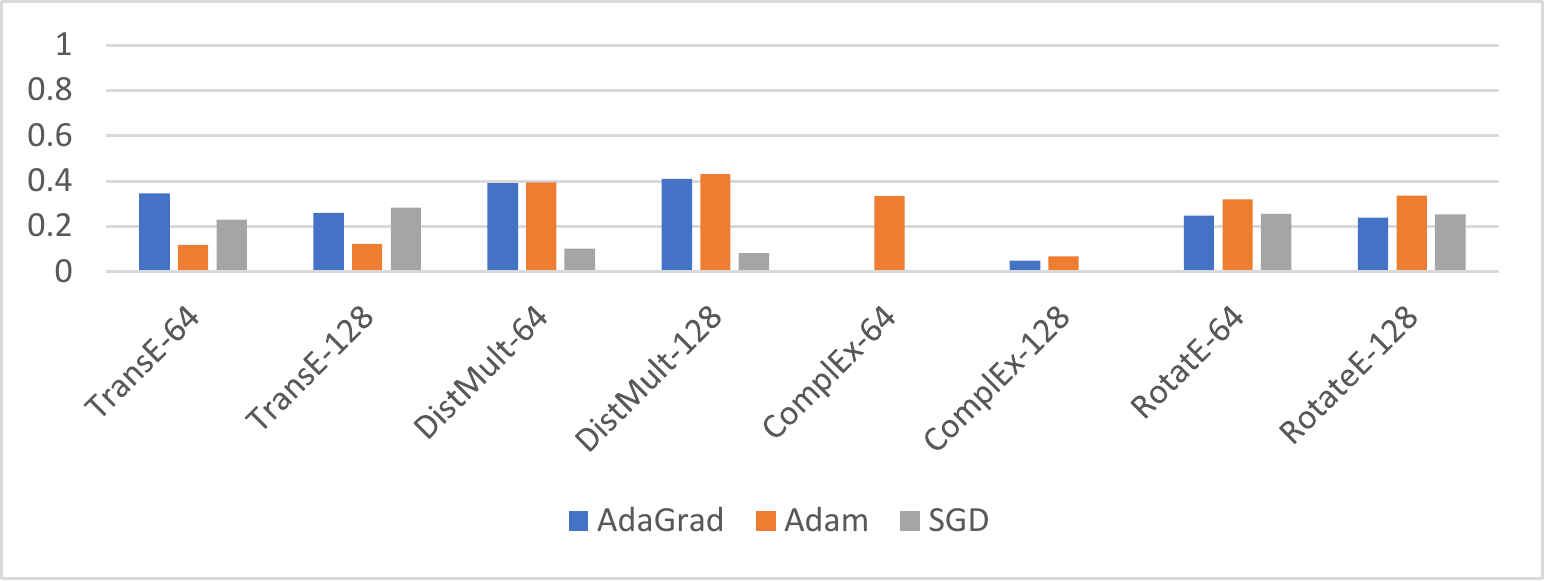}
        \caption{DBpedia50}
    \end{subfigure}
    \hfill
    \begin{subfigure}[t]{0.49\textwidth}
        \centering
        \includegraphics[width=\textwidth]{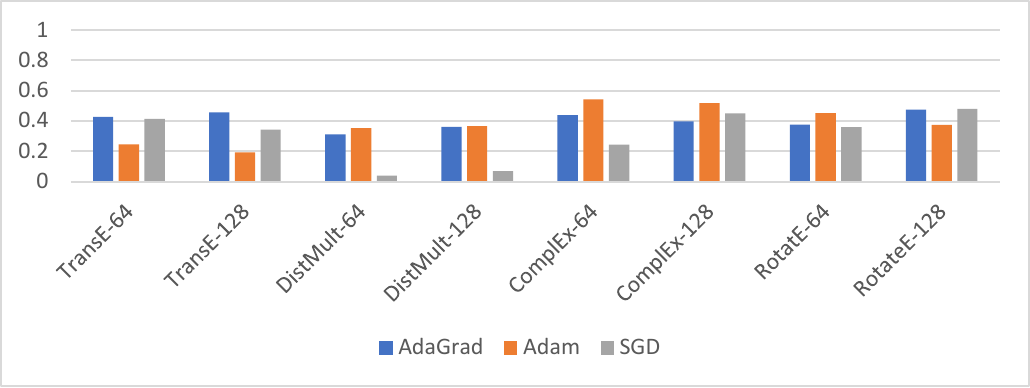}
        \caption{FB15k-237}
    \end{subfigure}
    \begin{subfigure}[t]{0.49\textwidth}
        \centering
        \includegraphics[width=\textwidth]{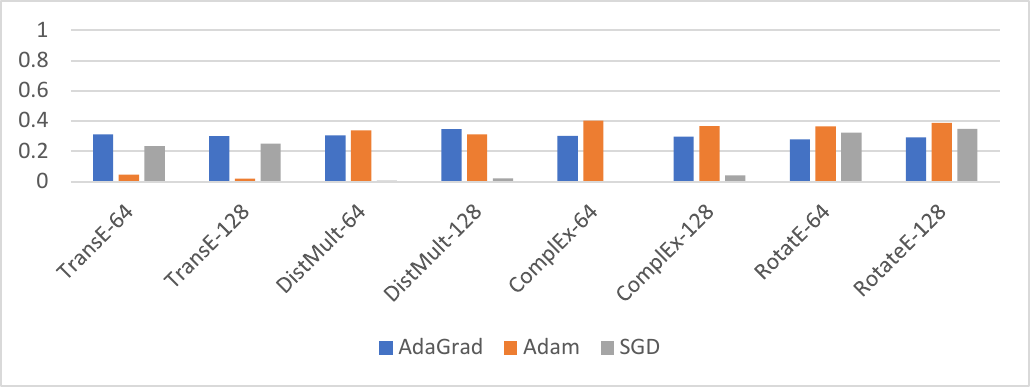}
        \caption{WN18RR}
    \end{subfigure}
    \hfill
    \begin{subfigure}[t]{0.49\textwidth}
        \centering
        \includegraphics[width=\textwidth]{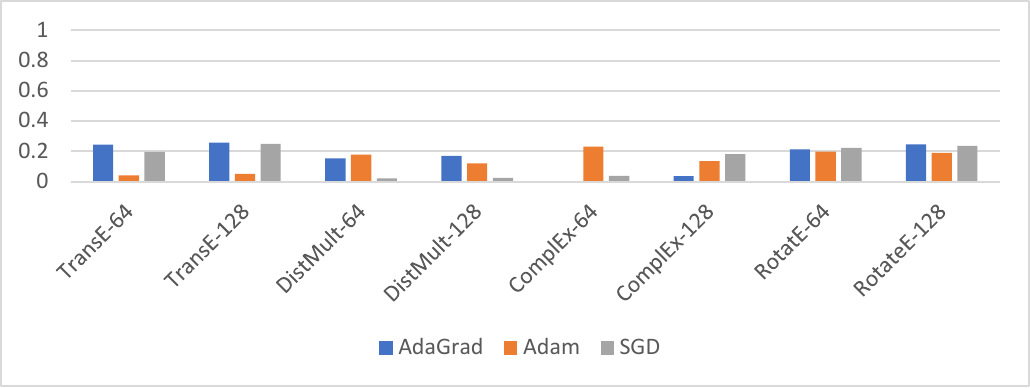}
        \caption{AristoV4}
    \end{subfigure}
    \caption{Ablation study on using different optimizers}
    \label{fig:ablation_optimizer}
\end{figure*}

Fig.~\ref{fig:ablation_optimizer} shows results obtained with varying the optimizer. There is quite a bit of variation between using different optimizers. It can be observed that, in most cases, the Adam optimizer works best with SnapE. Only when using SnapE for TransE (and ocassionally RotatE), AdaGrad and SGD are preferred.

\begin{figure*}[t]
    \centering
    \begin{subfigure}[t]{0.49\textwidth}
        \centering
        \includegraphics[width=\textwidth]{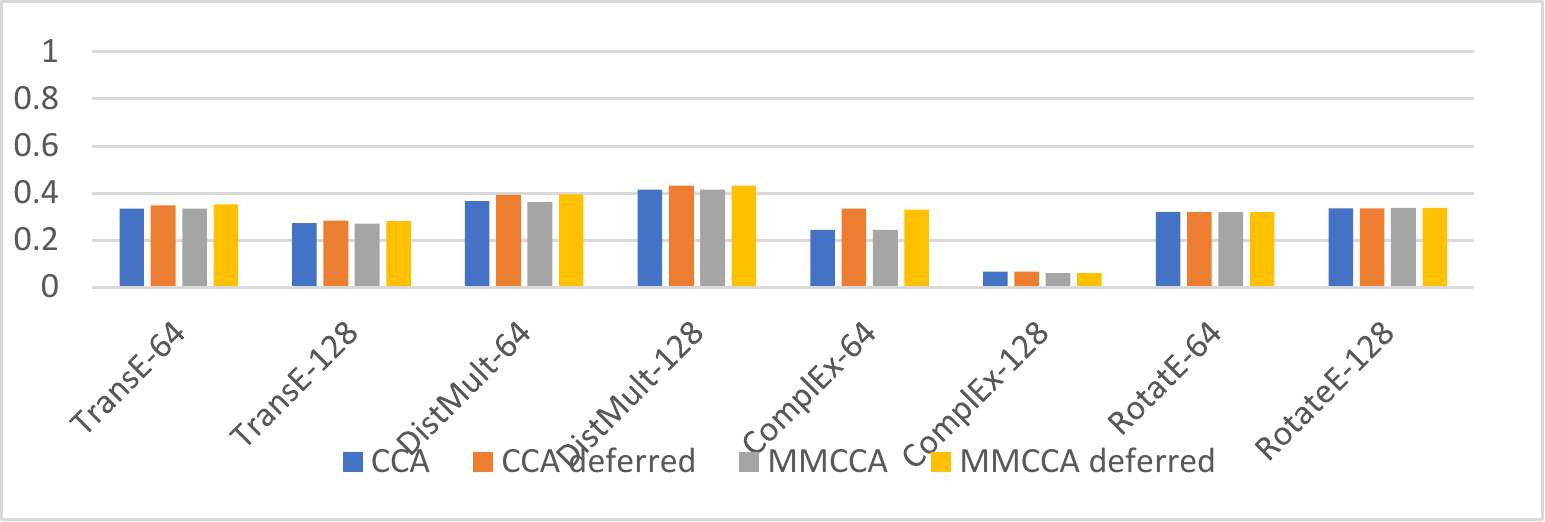}
        \caption{DBpedia50}
    \end{subfigure}
    \hfill
    \begin{subfigure}[t]{0.49\textwidth}
        \centering
        \includegraphics[width=\textwidth]{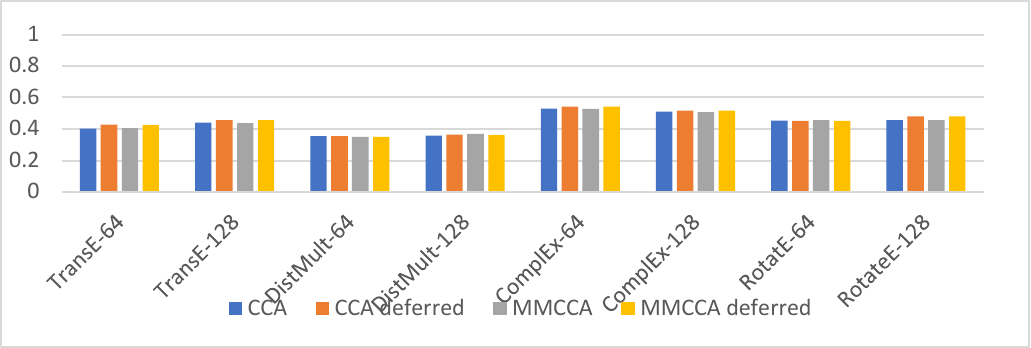}
        \caption{FB15k-237}
    \end{subfigure}
    \begin{subfigure}[t]{0.49\textwidth}
        \centering
        \includegraphics[width=\textwidth]{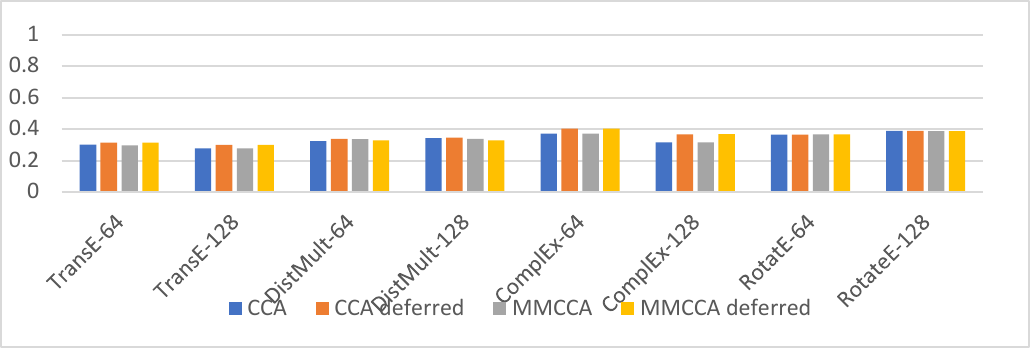}
        \caption{WN18RR}
    \end{subfigure}
    \hfill
    \begin{subfigure}[t]{0.49\textwidth}
        \centering
        \includegraphics[width=\textwidth]{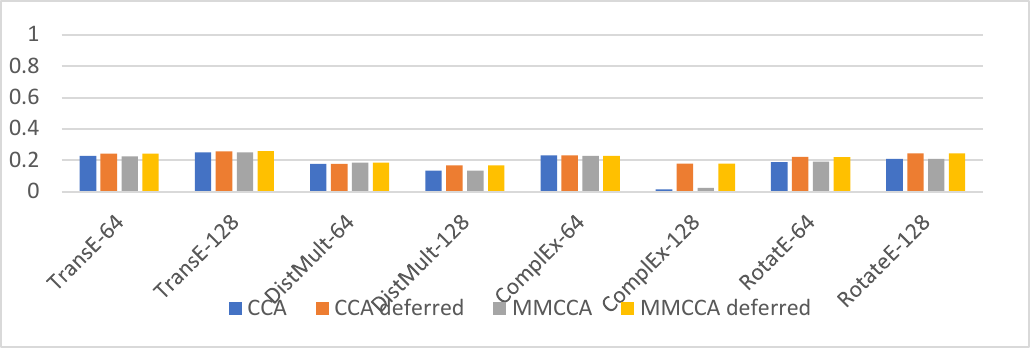}
        \caption{AristoV4}
    \end{subfigure}
    \caption{Ablation study on using different schedulers}
    \label{fig:ablation_scheduler}
\end{figure*}

Fig.~\ref{fig:ablation_scheduler} shows the results for using different scheduling mechanisms. It can be observed that in general, the influence of the scheduler is rather minimal. In cases where the results differ, the deferred variants have an advantage over the non-deferred variants (i.e., those scheduling variants may help, but rather do not hurt). There is only a small difference between standard CCA and MMCCLR. More detailed experiments have shown that CCA occasionally outperforms MMCCLR, but not vice versa.

\begin{figure*}[t]
    \centering
    \begin{subfigure}[t]{0.49\textwidth}
        \centering
        \includegraphics[width=\textwidth]{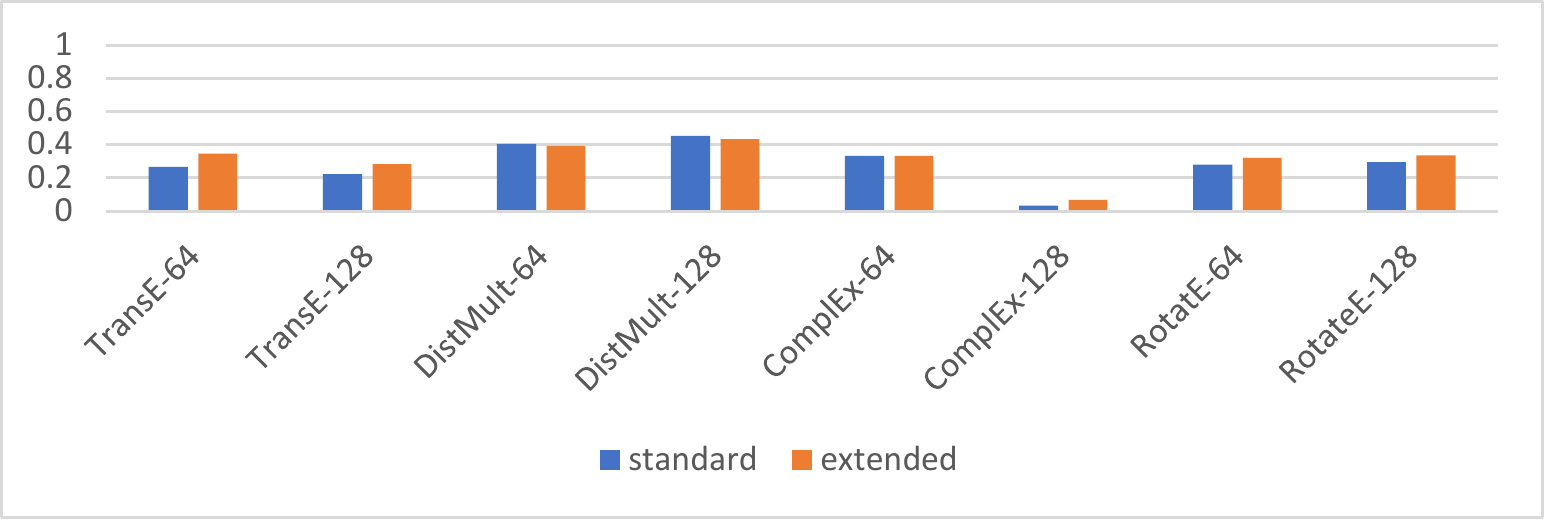}
        \caption{DBpedia50}
    \end{subfigure}
    \hfill
    \begin{subfigure}[t]{0.49\textwidth}
        \centering
        \includegraphics[width=\textwidth]{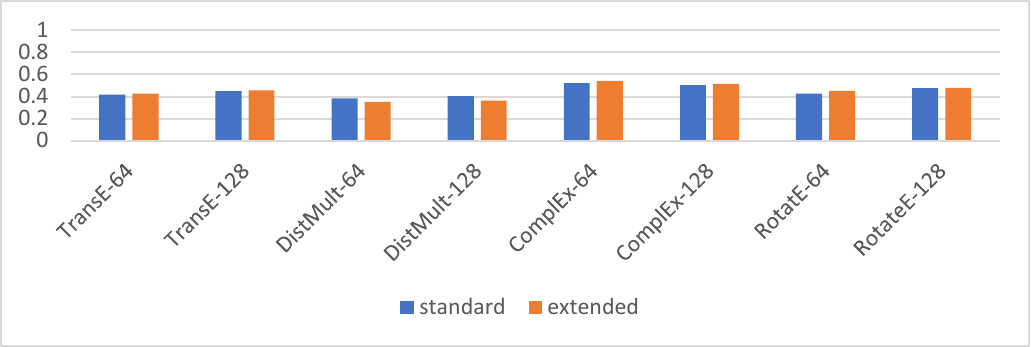}
        \caption{FB15k-237}
    \end{subfigure}
    \begin{subfigure}[t]{0.49\textwidth}
        \centering
        \includegraphics[width=\textwidth]{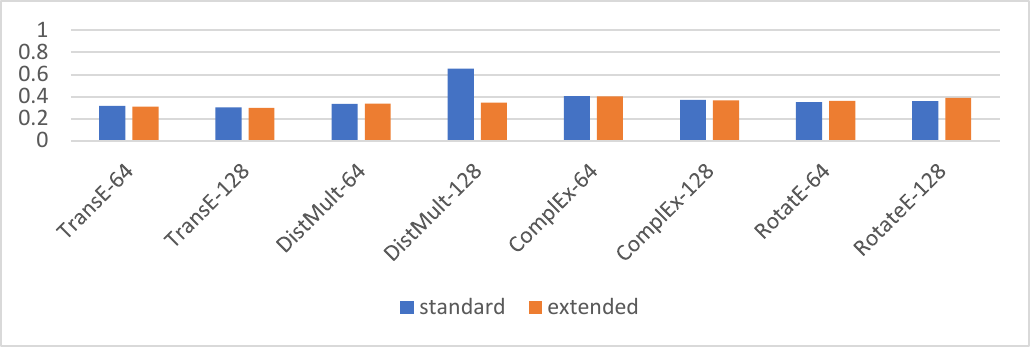}
        \caption{WN18RR}
    \end{subfigure}
    \hfill
    \begin{subfigure}[t]{0.49\textwidth}
        \centering
        \includegraphics[width=\textwidth]{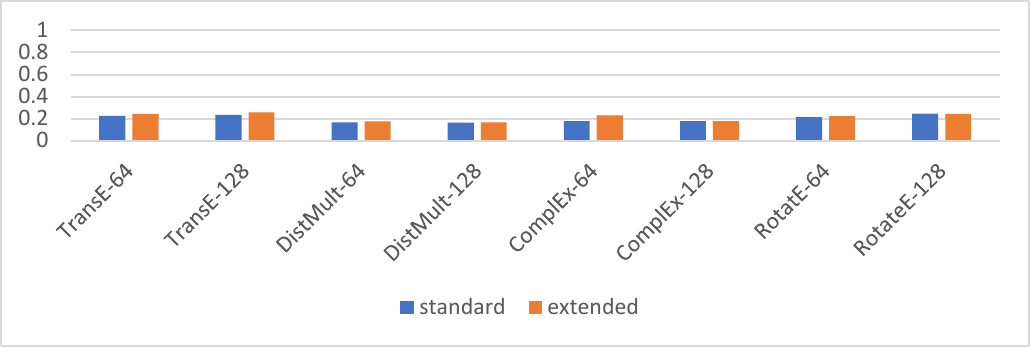}
        \caption{AristoV4}
    \end{subfigure}
    \caption{Ablation study on using the standard vs. the extended sampler}
    \label{fig:ablation_sampler}
\end{figure*}

Fig.~\ref{fig:ablation_sampler} shows the results obtained with standard negative sampling and the proposed extended negative sampler, which uses snapshots from previous iterations for negative sampling. It can be observed that the proposed extended negative sampler yields better results in most cases.
There is a clear outlier, which is the result for DistMult with d=128 on WN18RR. As shown in the analysis above, in that case, the variety of the models is very low when using the extended sampler (yielding a correlation of 0.967). This means that the individual models do not learn any different patterns (and, hence, the extended sampler cannot produce any interesting negatives for the subsequent iterations).


\begin{figure*}[t]
    \centering
    \begin{subfigure}[t]{0.49\textwidth}
        \centering
        \includegraphics[width=\textwidth]{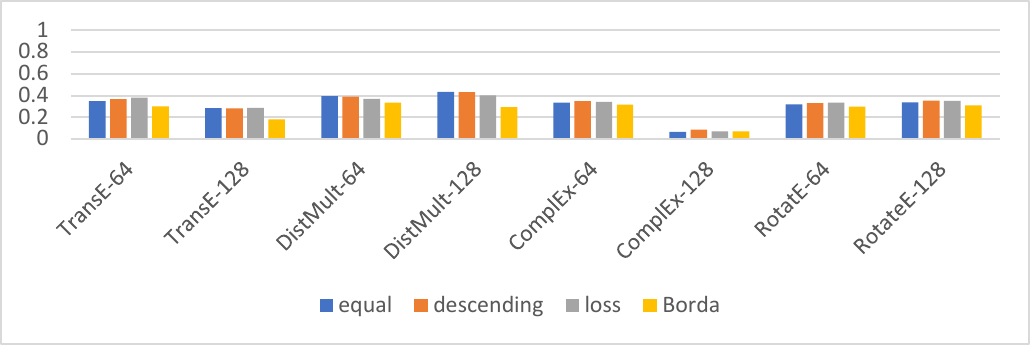}
        \caption{DBpedia50}
    \end{subfigure}
    \hfill
    \begin{subfigure}[t]{0.49\textwidth}
        \centering
        \includegraphics[width=\textwidth]{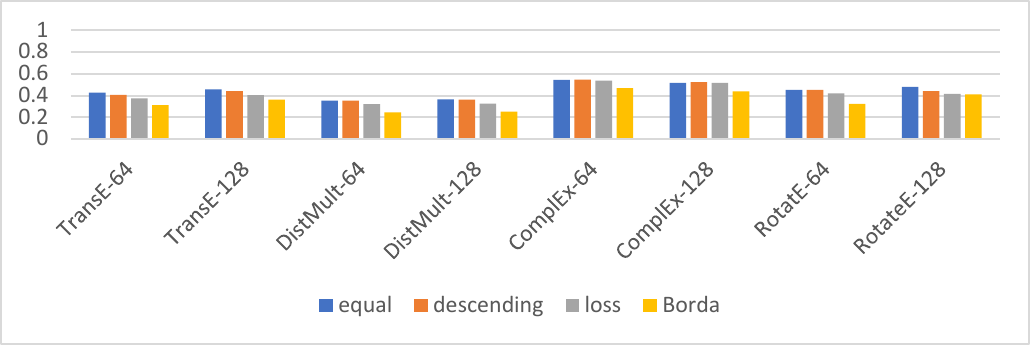}
        \caption{FB15k-237}
    \end{subfigure}
    \begin{subfigure}[t]{0.49\textwidth}
        \centering
        \includegraphics[width=\textwidth]{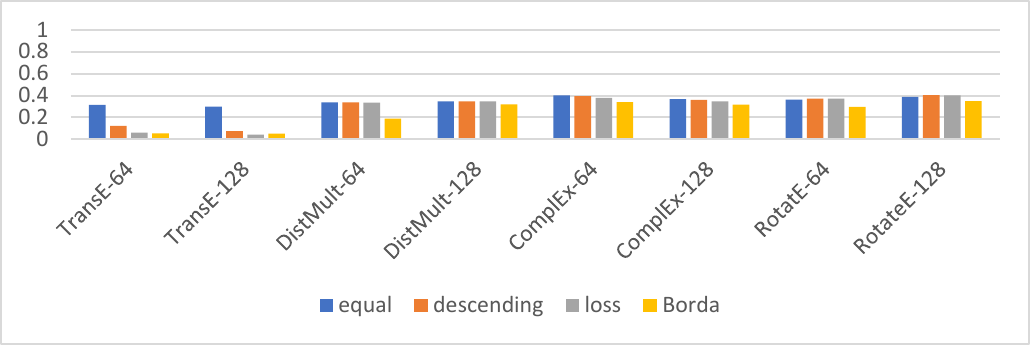}
        \caption{WN18RR}
    \end{subfigure}
    \hfill
    \begin{subfigure}[t]{0.49\textwidth}
        \centering
        \includegraphics[width=\textwidth]{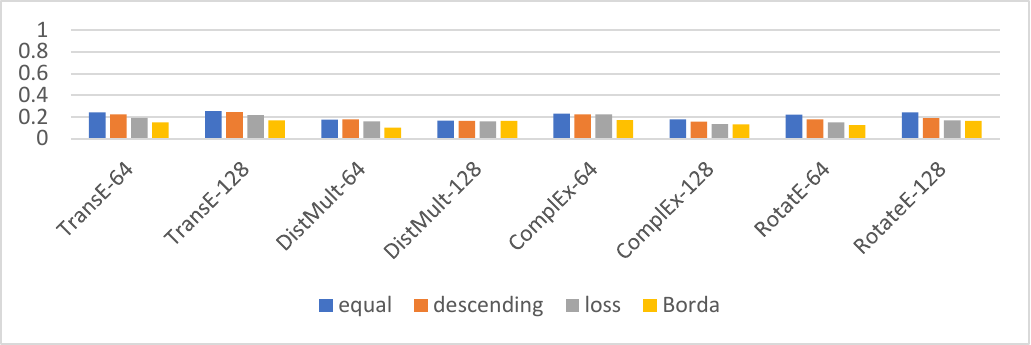}
        \caption{AristoV4}
    \end{subfigure}
    \caption{Ablation study on different strategies for combining predictions}
    \label{fig:ablation_combination}
\end{figure*}

Lastly, Fig.~\ref{fig:ablation_combination} shows a study of the impact of using different techniques for combining the results. We can observe that in general, the worst results are achieved with Borda rank aggregation, while the other methods are en par in many cases (with the exception of some cases with TransE, which work significantly better with equal weights). 

\section{Conclusion and Outlook}
\label{sec:conclusion}
In this paper, we have introduced SnapE, which is a training mechanism for knowledge graph embedding models based on snapshot ensembles. Instead of training one single model in $N$ epochs, SnapE trains a sequence of $C$ models with shorter training cycles of length $N/C$ each, using a decaying learning rate in each cycle. We have conducted experiments on four datasets, each time training SnapE with TransE, DistMult, ComplEx, and RotatE as base models. The resulting ensemble method has been shown to often deliver superior results to an individual model, without increasing the training cost.

The SnapE framework allows for further experimentation. We have observed that learning rate schedulers and combination approaches have an impact on the overall results, thus, we can assume that more experimentation with different variants here might yield even better results. As far as the learning rate schedulers go, early stopping could also be applied in each of the learning cycles, allowing for further optimization of the training time. The deferred training schedules could also allow to train the first iteration on a condensed version of the graph, as proposed in~\cite{hubert2023schema}.
For combining the predictions of snapshot models, techniques like mixture of experts~\cite{masoudnia2014mixture} could be a promising candidate for creating even more powerful snapshot ensembles.

While we could show that SnapE models can be trained at the same training cost as standard baseline models, we have observed a drastic increase in prediction time. At the same time, we could observe that it is often sufficient or even beneficial to only use a subset of the snapshots, which in turn lowers the prediction time. Thus, the selection strategy for the subset of snapshots seems a crucial property for further optimizing SnapE.

We have also observed that the possibilities that snapshot ensembles yield with respect to creating negative samples are intriguing. One could think of more elaborate techniques for exploiting negative samples from previous snapshots, e.g., mixing negative samples created by several previous snapshots.

Finally, it would be interesting to see how the idea of snapshot ensembles can be carried over to different kinds of embedding and graph learning methods, like R-GCNs~\cite{schlichtkrull2018modeling} or walk-based methods like RDF2vec~\cite{ristoski2016rdf2vec} and its variants~\cite{portischrdf2vec}, and other downstream tasks, such as node classification~\cite{bloem2021kgbench,pellegrino2020geval}. Moreover, they might be an interesting approach to embedding versioned knowledge graphs~\cite{hahnrdf2vec} by forming an evolving ensemble of snapshots trained on different versions.
\begin{table*}[t]
    \caption{Learning rate (lr), epochs (ep), and optimizer (Ad=Adam, AG=Adagrad) chosen for each base model as the best performing approach, based on HITS@10.}
    \label{tab:base_parameters}
    \scriptsize
    \centering
    \begin{tabular}{l||.|r|c||.|r|c||.|r|c||.|r|c}
         & \multicolumn{3}{c||}{DBpedia50} & \multicolumn{3}{c||}{FB15k237} & \multicolumn{3}{c||}{WN18RR} & \multicolumn{3}{c}{AristoV4} \\
         Model & LR & ep & O & LR & ep & O & LR & ep & O & LR & ep & O \\
         \hline
         TransE$_{d=64}$ & 1.0 & 110 & AG & 0.1 & 80 & AG & 0.1 & 220 & AG & 0.1 & 170 & AG \\
         TransE$_{d=128}$ & 1.0 & 80 & SGD & 0.1 & 120 & AG & 0.1 & 80 & AG & 0.1 & 150 & AG \\
         DistMult$_{d=64}$ & 0.1 & 90 & AG & 0.1 & 20 & AG & 0.1 & 10 & AG & 0.1 & 20 & AG \\
         DistMult$_{d=128}$ & 0.1 & 20 & AG & 0.1 & 20 & AG & 0.1 & 30 & AG & 0.01 & 400 & AG \\
         ComplEx$_{d=64}$ & 0.01 & 210 & Ad  & 1.0 & 350 & AG & 0.01 & 40 & Ad & 10 & 210 & SGD\\
         ComplEx$_{d=128}$ & 0.1 & 10 & Ad & 10 & 170 & SGD & 0.01 & 120 & Ad & 10 & 160 & SGD \\
         RotatE$_{d=64}$ & 10 & 100 & SGD & 1.0 & 270 & SGD & 10 & 70 & SGD & 0.1 & 490 & AG \\
         RotatE$_{d=128}$ & 10 & 110 & SGD & 1.0 & 400 & SGD & 10 & 80 & SGD & 0.1 & 150 & AG \\
    \end{tabular}
\end{table*}

\begin{table*}[t]
    \caption{Number of cycles (C), of models (M), and optimizer (O) for the best performing SnapE models, based on HITS@10.}
    \label{tab:snape_parameters}
    \scriptsize
    \centering
    \begin{tabular}{l||.|r|c||.|r|c||.|r|c||.|r|c}
         & \multicolumn{3}{c||}{DBpedia50} & \multicolumn{3}{c||}{FB15k237} & \multicolumn{3}{c||}{WN18RR} & \multicolumn{3}{c}{AristoV4} \\
         Model & C & M & O & C & M & O  & C & M & O  & C & M & O  \\
         \hline
         \multicolumn{13}{c}{SnapE -- Same Memory Budget}\\
         \hline
        TransE$_{d=64}$	&	10	&	2	&	Adagrad	&	10	&	2	&	Adagrad	&	10	&	2	&	Adagrad	&	10	&	2	&	Adagrad\\
        TransE$_{d=128}$	&	10	&	2	&	SGD	&	10	&	2	&	Adagrad	&	10	&	2	&	Adagrad	&	5	&	2	&	Adagrad\\
        DistMult$_{d=64}$	&	5	&	2	&	Adagrad	&	10	&	9	&	Adam	&	10	&	2	&	Adagrad	&	5	&	5	&	Adam\\
        DistMult$_{d=128}$	&	10	&	2	&	Adam	&	5	&	5	&	Adam	&	10	&	2	&	Adagrad	&	5	&	5	&	Adam\\
        ComplEx$_{d=64}$	&	10	&	2	&	Adam	&	10	&	2	&	Adagrad	&	10	&	2	&	Adam	&	5	&	2	&	Adam\\
        ComplEx$_{d=128}$	&	10	&	10	&	Adam	&	5	&	3	&	Adam	&	10	&	2	&	Adam	&	10	&	2	&	SGD\\
        RotatE$_{d=64}$	&	5	&	5	&	Adam	&	10	&	2	&	SGD	&	10	&	2	&	SGD	&	5	&	2	&	SGD\\
        RotatE$_{d=128}$	&	5	&	4	&	Adam	&	10	&	2	&	SGD	&	10	&	2	&	SGD	&	10	&	2	&	Adagrad\\

         \hline
         \multicolumn{13}{c}{SnapE -- Same Training Time Budget}\\
         \hline
        TransE$_{d=64}$	&	10	&	9	&	Adagrad	&	10	&	6	&	Adagrad	&	10	&	2	&	Adagrad	&	10	&	6	&	Adagrad\\
        TransE$_{d=128}$	&	10	&	8	&	SGD	&	10	&	6	&	Adagrad	&	10	&	2	&	Adagrad	&	10	&	8	&	Adagrad\\
        DistMult$_{d=64}$	&	10	&	3	&	Adam	&	10	&	10	&	Adam	&	10	&	9	&	Adam	&	10	&	10	&	Adam\\
        DistMult$_{d=128}$	&	10	&	6	&	Adam	&	10	&	9	&	Adam	&	10	&	3	&	Adagrad	&	10	&	2	&	Adagrad\\
        ComplEx$_{d=64}$	&	10	&	3	&	Adam	&	10	&	9	&	Adam	&	10	&	2	&	Adam	&	10	&	10	&	Adam\\
        ComplEx$_{d=128}$	&	10	&	10	&	Adam	&	10	&	7	&	Adam	&	10	&	2	&	Adam	&	10	&	2	&	SGD\\
        RotatE$_{d=64}$	&	10	&	10	&	Adam	&	10	&	9	&	Adam	&	10	&	10	&	Adam	&	10	&	4	&	SGD\\
        RotatE$_{d=128}$	&	10	&	10	&	Adam	&	10	&	2	&	SGD	&	10	&	10	&	Adam	&	10	&	2	&	Adagrad\\
    \end{tabular}
\end{table*}

\section*{Appendix: Configurations Used}
In this appendix, we show the configurations used for the baselines (table~\ref{tab:base_parameters}) as well as the SnapE models (table~\ref{tab:snape_parameters}) in section~\ref{sec:evaluation}.

\section*{Acknowledgements}
The authors acknowledge support by the state of Baden-Württemberg through bwHPC.
\bibliographystyle{splncs04}
\bibliography{thesis-references,additional_references}

\end{document}